\documentclass[11pt]{article}
\usepackage[final]{acl}
\usepackage{float}
\usepackage{stfloats}
\usepackage{placeins}
\usepackage{cuted}
\usepackage[hang, flushmargin]{footmisc}
\usepackage{comment}
\usepackage{times}
\usepackage{latexsym}
\usepackage[T1]{fontenc}
\usepackage[utf8]{inputenc}
\usepackage{microtype}
\usepackage{inconsolata}
\usepackage{graphicx}
\usepackage{amsmath}
\usepackage{amssymb}
\usepackage{multirow}
\usepackage{booktabs}
\usepackage{siunitx}

\usepackage[most]{tcolorbox}
\usepackage{algorithm}
\usepackage{algorithmic}
\usepackage{soul}
\usepackage{xcolor}
\definecolor{hlred}{RGB}{255,204,204}
\definecolor{hlgreen}{RGB}{204,255,204}
\definecolor{hlorange}{RGB}{255,226,179}
\definecolor{navy}{RGB}{35,85,145}
\definecolor{promptpurple}{RGB}{92,55,160}

\DeclareRobustCommand{\redhl}{\sethlcolor{hlred}\hl}
\DeclareRobustCommand{\greenhl}{\sethlcolor{hlgreen}\hl}

\newcommand{\promptplaceholder}[1]{\textcolor{promptpurple}{\{#1\}}}
\newtcolorbox{promptbox}[2][]{
   enhanced,
   unbreakable,
   colback=white,
   colframe=navy,
   colbacktitle=navy,
   coltitle=white,
   title={#2},
   fonttitle=\bfseries,
   toptitle=4pt,
   bottomtitle=4pt,
   lefttitle=10pt,
   righttitle=10pt,
   boxrule=1pt,
   arc=18pt,
   outer arc=18pt,
   width=0.96\linewidth,
   left=10pt,
   right=10pt,
   top=10pt,
   bottom=10pt,
   boxsep=0pt,
   before skip=8pt,
   after skip=16pt,
   halign=left,
   fontupper=\small\linespread{1.05}\selectfont,
   #1
}

\makeatletter
\newcommand{\beginpromptappendix}{%
   \begingroup
   \setlength{\textfloatsep}{6pt plus 1pt minus 1pt}%
   \setlength{\floatsep}{6pt plus 1pt minus 1pt}%
   \setlength{\dbltextfloatsep}{6pt plus 1pt minus 1pt}%
   \setlength{\dblfloatsep}{6pt plus 1pt minus 1pt}%
   \setlength{\@dblfptop}{0pt}%
   \setlength{\@dblfpsep}{6pt plus 1pt minus 1pt}%
   \setlength{\@dblfpbot}{0pt plus 1fil}%
   \captionsetup[figure]{skip=3pt}%
   \tcbset{promptcompact/.style={before skip=1pt,after skip=3pt,top=6pt,bottom=6pt,left=8pt,right=8pt,toptitle=2pt,bottomtitle=2pt,fontupper=\small\linespread{0.98}\selectfont}}%
}
\newcommand{\finishpromptappendix}{\endgroup}
\makeatother

\newtcolorbox{casebox}[1]{
   colback=orange!5,
   colframe=orange!40,
   title=\textbf{\textcolor{black}{#1}},
   coltitle=black,
   fonttitle=\bfseries,
   boxrule=1pt,
   arc=3mm,
   auto outer arc,
   left=6pt, right=6pt, top=4pt, bottom=4pt,
   titlerule=0pt,
   title filled=false
}
\title{Assembling the CREW: A Collaborative Multi-agent Reinforcement Learning Framework for Automated Related Work Generation}

\author{Hai-Dang Dang\textsuperscript{1*}, \quad Bao-Yen Pham\textsuperscript{1*}, \quad Bao Nguyen\textsuperscript{2}\\
         \textbf{Tran Thi Huong}\textsuperscript{3}, \quad \textbf{Huynh Thi Thanh Binh}\textsuperscript{1}\\
         \textsuperscript{1}Hanoi University of Science and Technology \quad
         \textsuperscript{2}The Chinese University of Hong Kong\\
         \textsuperscript{3}Hanoi University of Industry}

\begin{document}
\maketitle
\begingroup
\renewcommand{\thefootnote}{*}
\footnotetext[0]{Both authors contributed equally to this research.}
\endgroup
\begin{abstract}
Automatic Related Work Generation (RWG) significantly reduces the human time and effort required to author the Related Work Section (RWS) of a research paper. However, prior methods leveraging multi-agent Large Language Models (LLMs) typically rely on a predefined workflow, where each agent is responsible for a specific step in the entire process. This rigid, static inter-agent coordination limits the adaptive collaboration required to synthesize complex scientific literature. To address this limitation, we propose CREW (Collaborative Reinforcement Learning for Related Work Generation), a novel framework where LLM agents bypass heuristic pipelines to dynamically coordinate by autonomously selecting actions, such as Retrieve, Disseminate, Compose, and Critique, driven by a policy optimized via Independent Proximal Policy Optimization (IPPO). Extensive experiments on a standard RWG benchmark demonstrate that our approach yields substantial quality improvements over strong existing baselines, while significantly reducing token costs. Code is available at \url{https://github.com/YenPBao/CREW-Collaborative-MARL.git}
\end{abstract}

\section{Introduction}
To translate an initial idea into scientific research, scholars often explore a range of literature, carefully summarizing, synthesizing, and contrasting earlier findings. This process is both time-consuming and labor-intensive. Indeed, literature synthesis represents a major bottleneck in the scientific workflow, particularly given the rapid growth of scientific publications. Addressing this challenge, \citet{hoang-kan-2010-towards} introduced Related Work Generation (RWG) as a research direction within Natural Language Processing (NLP). The task aims to automatically transform an initial concept, typically consisting of a title and an abstract, together with a list of potential related references, into a coherent and logical related work draft that comprehensively covers key findings from prior studies with accurate citations.

Overall, RWG research has evolved along three main methodological directions.
Firstly, extractive approaches select and concatenate sentences directly from source documents, making them simple to implement and generally faithful to the original content. However, their limited ability to paraphrase and abstract often results in related-work sections that are redundant and poorly structured \citep{hu-wan-2014-automatic,chen2016summarization,deng2021automatic}. Secondly, abstractive supervised models move beyond sentence extraction by incorporating task-specific components and structured representations, including relation-aware RWG models and structured scientific summarization methods \citep{chen-etal-2021-capturing, chen2022target, xiao-etal-2022-primera}. Although these models achieve strong performance within their training domains, their reliance on task-specific training and annotated datasets limits scalability and adaptation to new domains or topics.
Thirdly, LLM-based methods leverage strong generative capabilities to produce coherent and logically structured related work sections. However, single-LLM approaches remain prone to hallucination and constrained by context capacity, as high-quality synthesis often requires jointly reasoning over many references. Even with long-context models (e.g., GPT-4o with a 128K-token capacity), simply concatenating all references into a single input remains ineffective, as long contexts can dilute attention and cause important inter-document relationships to be overlooked. Retrieval-based alternatives reduce the context burden but may omit relevant evidence and weaken coverage \citep{izacard-grave-2021-leveraging, wu2022memorizingtransformers}.
To overcome these obstacles, recent research has leveraged multi-agent LLM frameworks to decompose the overall RWG process into smaller subtasks, assigning each agent to a specific part of the workflow. This division of labor reduces the workload each agent must process, thereby lowering the risk of exceeding the context window \citep{wu2024autogen,li2023camel,liu-etal-2025-select}. However, fixed subtask assignments force the overall process to proceed sequentially, increasing inference time. More importantly, these approaches lack a mechanism that allows multiple agents to collaborate on the same subtask when it becomes critical. If too few agents are assigned to an important subtask at a particular stage, it can become a bottleneck, producing low-quality intermediate outputs that limit subsequent agents. For example, after agents have moved from reading to drafting, they may find that the collected evidence remains insufficient, making further drafting ineffective despite multiple agents being assigned to writing. Therefore, agents need mechanisms to proactively select context-dependent actions and coordinate more effectively.

Motivated by these observations, we introduce a multi-agent framework that enables flexible, context-dependent action selection through Multi-Agent Reinforcement Learning (MARL). MARL enables independently operating agents to coordinate their decisions through a shared learned policy rather than a centralized orchestrator, thereby reducing computational overhead and improving scalability. To achieve this, we formulate RWG as a Decentralized Partially Observable Markov Decision Process (Dec-POMDP), where each agent receives a distinct local observation consisting of its internal knowledge state, shared resources, and peer actions. At the same time, our environment provides a team-level global reward based on the quality of the collaboratively constructed draft and its alignment with human-written references. Importantly, beyond the conventional actions of literature retrieval and drafting employed in prior work, we introduce two additional actions, \emph{knowledge dissemination} and \emph{draft critiquing}, to promote proactive information sharing and adaptive inter-agent coordination throughout the RWG process. The framework is then optimized using IPPO with parameter sharing to ensure computational efficiency and scalability. Extensive experiments show that our framework improves over competitive baselines in LLM-as-a-judge evaluation and valid citation grounding, while reducing inference time and token costs.

Our contributions can be summarized as follows:

\begin{enumerate}

  \item We introduce two collaborative actions, Disseminate and Critique, which enable agents to share useful knowledge and provide mutual feedback, improving both coordination and the quality of the generated draft.

  \item We propose a MARL framework based on a Dec-POMDP formulation, allowing agents to proactively select context-dependent actions from their local observations. We further optimize the decision-making policy using IPPO with parameter sharing for efficient multi-agent coordination.

  \item We conduct extensive experiments showing that our framework improves over strong existing methods across multiple evaluation metrics. Additional transferability experiments demonstrate that a decision-making policy trained with one LLM can be reused with different inference LLMs, indicating that the learned coordination strategy is not tied to a single backbone model.

\end{enumerate}

\section{Related Work}

RWG has attracted growing attention for its role in supporting scientific writing and is commonly categorized into Extractive RWG (ERWG) and Abstractive RWG (ARWG)~\citep{li-ouyang-2024-related}. ERWG methods construct related work sections by selecting and concatenating representative sentences from cited papers. Early approaches such as ReWoS~\citep{hoang-kan-2010-towards} rely on hierarchical rule-based extraction at both coarse- and fine-grained topic levels, while subsequent methods incorporate topic modeling and learning-based importance estimation, e.g., clustering sentences with PLSA and ranking them using regression models~\citep{hu-wan-2014-automatic}. To better capture target–reference relationships, later studies exploit citation contexts~\citep{chen2016summarization} or sentence-level embeddings with coherence-aware reordering, as in SERGE~\citep{deng2021automatic}. Graph-based methods explicitly model inter-paper relations, such as ToC-RWG~\citep{8931592} and bibliographic graph-based neural summarization~\citep{wang-etal-2018-neural-related}. Overall, despite being citation-faithful, ERWG methods remain limited in producing comparative and analytical discussions, as they depend heavily on pre-existing comparative sentences that are difficult to obtain in practice.

To address these limitations, recent work has shifted toward ARWG, where models generate related-work text in their own words. ARWG approaches can be broadly categorized into citation-level and section-level generation. Citation-level ARWG focuses on generating short descriptions for individual citations based on local citation context and cited paper content, typically using citation context and abstracts due to input length constraints~\citep{luu-etal-2021-explaining,gu2024controllable, jung2022intent, li2024cited, xiao-etal-2022-primera}. These methods generally assume one cited paper per citation and model citations in isolation, limiting their ability to capture cross-paper relationships. Section-level ARWG extends this formulation by generating longer related-work paragraphs that contain multiple citations, explicitly modeling interactions among referenced papers to improve global coherence. Following early attention-based models~\citep{wang-etal-2018-neural-related}, a common strategy is to model papers and their interactions as graphs, enabling structured reasoning over citations, topics, and semantic dependencies. Building upon this paradigm, \citet{chen-etal-2021-capturing} introduced RRG, which captures relationships among referenced papers using a relation graph and iteratively refines document representations prior to generation. Similarly, \citet{chen2022target} proposed TAG, integrating graph-guided interactions at both the encoding and decoding stages, alongside contrastive learning, to encourage well-structured outputs. In this work, we focus on section-level ARWG.

Despite their demonstrated effectiveness, existing section-level ARWG methods are typically trained from scratch, resulting in substantial computational and memory overhead when scaled to large corpora with many reference papers. To overcome this, a promising research direction is to leverage the robust generative capabilities of pretrained LLMs to bypass costly retraining. However, fully unlocking LLMs for this task requires overcoming their inherent context-window limitations. This challenge is especially pronounced in section-level ARWG, where identifying complex latent links necessitates simultaneously processing a massive volume of text across numerous reference papers. Prior attempts to handle such input-length constraints commonly rely on long-context models, iterative reading-and-summarization strategies~\citep{miller-etal-2016-key,chevalier-etal-2023-adapting}, and retrieval-based methods~\citep{izacard-grave-2021-leveraging, wu2022memorizingtransformers}. However, each direction has notable drawbacks: long-context processing can dilute the effectiveness of attention, iterative reading may overwrite early evidence, and retrieval risks omitting crucial inter-document nuances.

To circumvent these limitations, our framework leverages multi-LLM collaboration to effectively distribute reasoning and knowledge processing across multiple agents. Unlike many existing multi-LLM systems that rely on rigid predefined interaction patterns or static self-reflection loops~\citep{wu2024autogen, li2023camel, shinn2023reflexion}, our approach proposes a highly adaptable paradigm grounded in multi-agent reinforcement learning. For instance, while \citet{liu-etal-2025-select} recently proposed a multi-agent system that uses citation graphs and specialized roles (Selector, Reader, and Writer), imposing these fixed roles forces agents into sequential pipelines, restricting their ability to address subtasks as they become critical. Conversely, our approach offers a flexible collaboration paradigm by introducing explicit dynamic cooperation via discussion actions among complementary LLM agents and enabling step-wise situational action selection, which allows agents to dynamically determine subsequent actions, reducing unnecessary interactions and computational overhead while improving reasoning reliability.

\begin{figure*}[t] 
    \centering
    \includegraphics[width=1.0\textwidth]{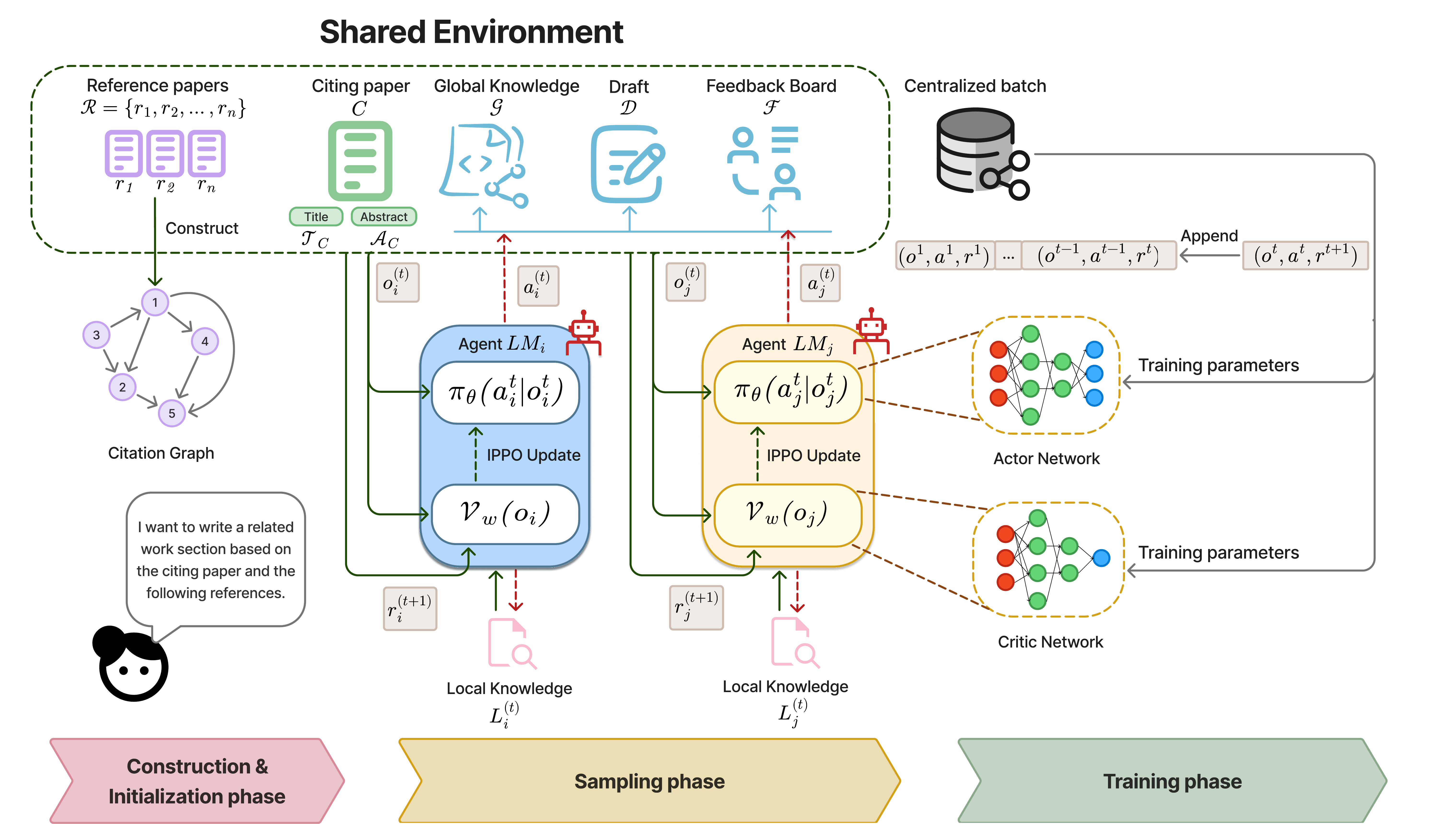} 
    \caption{Overview of our multi-agent framework. Agents operate independently and in parallel over a series of timesteps. At each step, each agent observes the input information and current working documents, selects one of four actions, and receives a reward guided by the human-written related-work section as the gold standard. The collected interactions are aggregated into a buffer to train the policy, which is then used as the agents' decision-making policy during inference to produce the final output.}
    \label{fig:system_architecture}
\end{figure*}
\section{Methodology}
\label{sec::environment}
\noindent Following the setting of Select, Read, and Write~\citep{liu-etal-2025-select}, the input consists of the author's initial idea, represented by the title $\mathcal{T}_C$ and abstract $\mathcal{A}_C$ of the citing paper $C$, together with a set of potential reference papers $\mathcal{R}=\{r_1,r_2,\dots,r_n\}$. Given these inputs, our framework transforms the reference set into a coherent related work section $\mathcal{S}$ that synthesizes the key ideas most relevant to $C$ with accurate citation grounding. The overall workflow of the proposed framework is summarized in Figure~\ref{fig:system_architecture}.

\subsection{Dec-POMDP Formulation}
\label{sec:dec-posmdp}
In this work, we cast RWG as a Dec-POMDP, which naturally accommodates
(i)~\emph{partial observability}, since each of $M$ agent sees only the shared working documents and its own private knowledge, (ii)~all agents act \emph{independently} based on their own distinct observations, without relying on a centralized controller, thereby distributing the computation and enhancing scalability. Thus, agents operate independently and in parallel over a series of timesteps: at each timestep~$t$, the environment is in state~$s_t$, each agent~$i$ receives an observation~$o_t^i$ as its partial view of the state, selects an action~$a_t^i$, and then receives a reward~$r_t^i$ from the environment. We define the \emph{states}, \emph{observations}, \emph{actions}, and \emph{rewards} as follows.

\textbf{States.}
At each time step~$t$, the global state $s_t\in\mathcal{S}$ is defined as:
\begin{equation}
\label{eq:state}
s_t = \langle \mathcal{T}_C, \mathcal{A}_C, \mathcal{R}, L_t, G_t, D_t, F_t \rangle,
\end{equation}
where the first three components are the static task inputs introduced above. The remaining four components correspond to the four types of documents that we propose for the collaborative writing process.
\emph{\textbf{Local knowledge}} $L_t=\{L_t^{1},\ldots,L_t^{M}\}$ is the collection of agent-private knowledge states; each $L_t^{i}$ accumulates reading summaries obtained through Agent~$i$'s individual exploration of the citation graph via the Retrieve action and remains \emph{invisible} to other agents. \emph{\textbf{Global knowledge}} $G_t$ is a shared knowledge base accessible to all agents, aggregated from individual local contributions via the Disseminate action; it serves as the collective understanding of the literature that any agent can draw upon when drafting or critiquing.
\emph{\textbf{Draft}} $D_t$ is the evolving text of the related-work section, iteratively refined by agents through the Compose action.
\emph{\textbf{Feedback}} $F_t$ stores the accumulated critiques and revision suggestions produced by agents through the Critique action, thereby providing structured signals to systematically guide subsequent draft revisions. The detailed explanation of the \emph{Actions} is provided below.

\textbf{Observations.}
Since Agent~$i$ cannot access the local knowledge of other agents, each agent receives only a partial observation $o_t^i\in\mathcal{O}^i$:
\begin{equation}
\label{eq:obs}
o_t^{i} = \mathrm{Embed}\!\left(\bigl\langle\; L_t^{i},\; G_t,\; D_t,\; F_t \;\bigr\rangle\right),
\end{equation}
Directly feeding the raw textual content of each document into the policy network would result in an excessively large observation space, making convergence during training prohibitively difficult. To maintain a compact state representation, the embedding operator $\mathrm{Embed}(\cdot)$ encodes each component $k\in\{L,G,D,F\}$ into a five-dimensional numerical feature vector $\mathbf{x}_k = [v_{1,k},\ldots,v_{5,k}]$, with all feature values clipped to $[0,1]$. The resulting observation has 20 dimensions and is independent of the number of agents~$M$, so the policy architecture does not need to change when the team size varies. The five features are defined as follows:

\noindent(i)~\emph{\textbf{Semantic relevance}} ($v_{1,k}$): The cosine similarity, denoted by $c(\cdot,\cdot)$, between the embedding $\mathbf{e}_k$ of component~$k$ and the initial-idea embedding $\mathbf{e}_{\mathrm{idea}}=\mathrm{embed}(\mathcal{T}_C \oplus \mathcal{A}_C)$, where the initial idea consists of the title $\mathcal{T}_C$ and abstract $\mathcal{A}_C$:
\begin{equation}
\label{eq:obs_semantic_relevance}
v_{1,k} = c(\mathbf{e}_k, \mathbf{e}_{\mathrm{idea}}).
\end{equation}

\noindent(ii)~\emph{\textbf{Coverage and Novelty}} ($v_{2,k}$): A component-specific measure defined as:
\begin{equation}
\label{eq:obs_coverage_novelty}
v_{2,k}\!=\!
\begin{cases}
1 - \alpha\,\delta_{\mathrm{id}} - (1\!-\!\alpha)\,\delta_{\mathrm{sem}}\!, & k\!=\!L,\\[2pt]
1-c(\mathbf{e}_{L_i},\mathbf{e}_G), & k\!=\!G,\\[2pt]
c(\mathbf{e}_D,\mathbf{e}_G), & k\!=\!D,\\[2pt]
c(\mathbf{e}_F,\mathbf{e}_{\mathrm{idea}}\!-\!\mathbf{e}_D), & k\!=\!F,
\end{cases}
\end{equation}
where for Local Knowledge ($L$), the coverage feature penalizes redundancy against other agents. Specifically, $\delta_{\mathrm{id}} = |\mathcal{P}_i \cap \mathcal{P}_{\neg i}| / |\mathcal{P}_i|$ calculates the identifier-level overlap ratio between the set of papers read by Agent~$i$ ($\mathcal{P}_i$) and those read by all other agents ($\mathcal{P}_{\neg i}$). Concurrently, $\delta_{\mathrm{sem}} = \max_{j\neq i} c(\mathbf{e}_{L_i}, \mathbf{e}_{L_j})$ captures the maximum semantic similarity between Agent~$i$'s local knowledge and that of any peer. For Global Knowledge ($G$), it measures semantic novelty relative to the shared knowledge. For the Draft ($D$), it quantifies semantic coverage with reward to the $G$. For Feedback Board $F$, it measures alignment with the \emph{gap vector} $\mathbf{e}_{\mathrm{idea}}-\mathbf{e}_D$.

\noindent(iii)~\emph{\textbf{Personal action distribution}} ($v_{3,k}$): The historical frequency with which Agent~$i$ has selected action~$k$, defined explicitly as:
\begin{equation}
\label{eq:obs_personal_action_distribution}
v_{3,k} = \frac{n_k^i}{\sum_{a\in\mathcal{A}} n_a^i},
\end{equation}
where $n_k^i$ is the accumulated count of action~$k$ executed by Agent~$i$.

\noindent(iv)~\emph{\textbf{Token capacity}} ($v_{4,k}$): The remaining normalized token capacity for action~$k$:
\begin{equation}
\label{eq:obs_token_capacity}
v_{4,k} = \max\!\left(0, 1 - \frac{\tau_k}{\tau_{\max}}\right),
\end{equation}
where $\tau_k$ is the current token usage associated with action~$k$, and $\tau_{\max}$ represents the predefined maximum allowable token budget.

\noindent(v)~\emph{\textbf{Social activity distribution}} ($v_{5,k}$): The proportion of action~$k$ performed by \emph{other} agents,
\begin{equation}
\label{eq:obs_social_activity_distribution}
v_{5,k} = \frac{n_k^{\neg i}}{n_k^{\mathrm{all}}},
\end{equation}
providing Agent~$i$ with awareness of team-level coordination, where $n_k^{\neg i}$ is the number of times action~$k$ has been performed by all agents except Agent~$i$, and $n_k^{\mathrm{all}}$ is the total number of times action~$k$ has been performed by all agents.

\textbf{Actions.}
The joint action space is $\mathcal{A}=\mathcal{A}^1\times\cdots\times\mathcal{A}^M$; each agent~$i$ selects one of four actions: Retrieve, Disseminate, Compose, or Critique, and the joint action is $\mathbf{a}=(a^1_t,\ldots,a^M_t)$.
Each action is realized by a dedicated prompt template (Appendix~\ref{sec:appendix_prompts}).
The Retrieve action selects a candidate paper $p_c^{t-1}$ to read and updates the agent's local knowledge.
Disseminate merges local into shared knowledge;
Critique reviews the current draft;
Compose revises the draft from all available sources:
\begin{align}
\label{eq:action_updates}
L_t^{i} &= \mathrm{Retrieve}\bigl(L_{t-1}^{i},\, p_c^{t-1}\bigr) \\
G_t &= \mathrm{Disseminate}\bigl(L_t^{i},\, G_{t-1}\bigr) \\
F_t &= \mathrm{Critique}\bigl(L_t^{i},\, D_{t-1}\bigr) \\
D_t &= \mathrm{Compose}\bigl(D_{t-1},\, G_{t-1},\, L_t^{i},\, F_{t-1}\bigr)
\end{align}

\textbf{Rewards.}
We adopt a \emph{delta-based} cooperative reward structure. The reward decomposes into two terms: a \emph{draft-quality gain} reflecting overall synthesis progress, and a \emph{component-quality gain} reflecting improvements in the specific artifact targeted by the agent's action:
\begin{equation}
\label{eq:reward}
r_t^{i} = (1-\lambda^t)\,\Delta Q_D^{(t)} + \lambda^t\,\Delta Q_{k(a^i)}^{(t)},
\end{equation}
where $Q_k^{(t)}$ denotes the quality of component~$k\in\{L,G,D,F\}$ at time step~$t$, and $\Delta Q_k^{(t)}=Q_k^{(t)}-Q_k^{(t-1)}$ denotes its quality gain. The decaying coefficient $\lambda^t$ encourages agents to first improve the background documents before gradually shifting attention toward composing the final draft.
The quality $Q_k$ is estimated by document-specific heuristic functions, depending on the type of document. These heuristics primarily reference the human-written related-work section, whose embedding is denoted by $\mathbf{e}_{\mathrm{gold}}$. Detailed quality formulas and the credit-assignment procedure are provided in Appendix~\ref{app:reward} and Appendix~\ref{app:credit_assignment}.

\subsection{Proposed Algorithm}

As illustrated in Figure~\ref{fig:system_architecture}, we optimize the cooperative policy using parameter-shared IPPO, with full pseudocode provided in Appendix~\ref{sec:appendix_A}. The framework alternates between sampling and training: agents interact asynchronously under a shared behavioral policy, collect rewards from Section~\ref{sec:dec-posmdp}, and aggregate rollouts to update a single shared Actor and Critic. This design keeps the policy scalable while reducing coordination overhead across agents.

\textbf{The sampling phase.}
During data collection, $M$ agents asynchronously select actions from the discrete space $\mathcal{A}$ based on the shared policy $\pi_\theta$. This interaction occurs over a subset of papers, spanning several time steps across multiple epochs. The reward is calculated by measuring the specific semantic and structural improvements made to the draft, compared with the ground truth one. These transition tuples $(o_t^i, a_t^i, r_t^i, o_{t+1}^i)$ are temporarily stored in local experience buffers.

\textbf{The training phase.}
Once a predefined stopping condition is met, trajectories from all local buffers are aggregated into a centralized batch $\mathcal{W}$ to update the shared networks. For each sample $j\in\mathcal{W}$, PPO defines the probability ratio
\begin{equation}
\label{eq:ppo_ratio}
\rho_j(\theta)=\frac{\pi_{\theta}(a_j\mid o_j)}{\pi_{\theta_{\mathrm{old}}}(a_j\mid o_j)}.
\end{equation}
The clipped surrogate objective for the Actor is
\begin{multline}
\label{eq:ppo_clip_objective}
\mathcal{L}^{\mathrm{CLIP}}(\theta)=
\frac{1}{|\mathcal{W}|}\sum_{j\in\mathcal{W}}
\min\Bigl(\rho_j(\theta)\hat{A}_j, \\
\mathrm{clip}(\rho_j(\theta),1-\epsilon,1+\epsilon)\hat{A}_j\Bigr),
\end{multline}
where $\hat{A}_j$ is estimated with Generalized Advantage Estimation (GAE)~\citep{schulman2016high}. The Critic is trained against the empirical return $\hat{R}_j$ using the value loss
\begin{equation}
\label{eq:value_loss}
\mathcal{L}^{\mathrm{VF}}(\phi)=
\frac{1}{|\mathcal{W}|}\sum_{j\in\mathcal{W}}
\left(V_{\phi}(o_j)-\hat{R}_j\right)^2.
\end{equation}
During optimization, the Actor maximizes the clipped surrogate objective while the Critic minimizes the value loss; this centralized optimization cycle is repeated until the policy converges.

\textbf{The actor and critic networks.}
To ensure parameter efficiency and facilitate knowledge transfer across agents, our framework maintains a single Actor network ($\pi_\theta$) and Critic network ($V_\phi$) applied uniformly to all agents.
The Actor maps the 20-dimensional observation vector to action logits $\mathbf{z} \in \mathbb{R}^{4}$ for categorical sampling, while the Critic outputs a scalar value estimate $\hat{V}_\phi(o_t^i)$ as the variance-reduction baseline. Both networks share the same compact feedforward design, differing only in their final output heads; detailed configurations are provided in Table~\ref{tab:system_parameters} in Appendix~\ref{sec:appendix_d}.

\section{Experimental Results and Discussions}
\setcounter{footnote}{0}
To evaluate the CREW framework, we design a series of experiments addressing overall performance, architectural choices, and cost efficiency.

\subsection{Experimental Setup}
\label{sec:exp_setup}

\textbf{Dataset.} We evaluate our framework on the \textsc{OARelatedWork} dataset~\citep{docekal2024oarelatedwork}, currently the primary dataset supporting full-text-based related work generation. Specifically, we use 2\% of the papers (1829 papers) for training and 0.4\% of the papers (366 papers) for evaluation, as detailed in Appendix~\ref{sec:appendix_b}.

\textbf{Implementation Details.} For the main results, we use Qwen 2.5 72B Instruct\footnote{Version qwen-2.5-72b-september2024\label{fn:qwen725}} for both policy training and inference. In the transferability study, we keep the trained controller fixed and swap the inference backbone to LLaMA 8B\footnote{Version llama-3.1-8b-instruct-july2024\label{fn:llama8b}}, Gemini 2.0 Flash\footnote{Version gemini-2.0-flash-february2025\label{fn:gemini20}}, Qwen 2.5 72B Instruct, and GPT-4o\footnote{Version gpt-4o-may2024\label{fn:gpt4o}}. Complete IPPO hyperparameters and network architecture are provided in Appendix~\ref{sec:appendix_d}, while prompt templates are provided in Appendix~\ref{sec:appendix_prompts}.

\textbf{Evaluation Metrics.} Following the evaluation protocol of Select, Read, and Write (SRW)~\citep{liu-etal-2025-select}, we adopt an \textit{LLM-as-a-Judge} setup. For robustness and fairness, final scores are computed as the arithmetic mean of ratings from three judges on a 1--5 Likert scale: GPT-4o, LLaMA 3.3 70B Instruct\footnote{Version llama-3.3-70b-december2024\label{fn:llama33}}, and DeepSeek-V3\footnote{Version deepseek-v3-december2024\label{fn:deepseekv3}}. In addition to standard metrics (\textit{Coverage}, \textit{Logic}, and \textit{Relevance}), we report \textit{Citation Verification}, defined as the proportion of generated citations that exist in the database. To assess operational efficiency, we further report total input/output tokens and inference time for the baselines. Detailed definitions and evaluation procedures for all metrics are provided in Appendix~\ref{sec:appendix_c}.

\subsection{Baselines}
\label{sec:exp_baselines}
We situate our proposed CREW framework against widely-adopted baseline paradigms (with detailed behavioral analysis in Appendix~\ref{sec:appendix_d3}):

\textbf{Multi-Agent Methods.} We compare against SRW~\citep{liu-etal-2025-select} and our own Orchestrator baseline, inspired by HuggingGPT~\citep{shen2023hugginggpt}, with both baselines using Qwen 2.5 72B Instruct to match the main setting. The Orchestrator incorporates CREW's proposed architecture, including the shared workspace, local knowledge, and action space; its key distinction is that it replaces the autonomous RL-based policy controller with a centralized LLM coordinator that assigns actions to the agents. Further implementation and behavioral details are provided in Appendix~\ref{sec:appendix_d3}.

\textbf{Single LLM Methods.} We evaluate GPT-4o with either the full document context or standard Retrieval-Augmented Generation (RAG)~\citep{lewis-etal-2020-retrieval}.

\textbf{Abstractive Methods.} We benchmark against PRIMERA~\citep{xiao-etal-2022-primera}, a strong sequence-to-sequence model for multi-document synthesis. 

\begin{table*}[!t]
\centering
\small
\setlength{\tabcolsep}{4pt}
\caption{Performance of different models on the \textsc{OARelatedWork} dataset. The best and the runner-up results are shown in \textbf{bold} and \underline{underlined}, respectively. Our method outperforms the strong existing baseline across all quality and citation metrics while using fewer tokens.}
\label{tab:main_results}
\begin{tabular*}{\textwidth}{@{\extracolsep{\fill}}>{\raggedright\arraybackslash}p{0.28\textwidth} >{\centering\arraybackslash}p{0.07\textwidth} >{\centering\arraybackslash}p{0.07\textwidth} >{\centering\arraybackslash}p{0.07\textwidth} >{\centering\arraybackslash}p{0.08\textwidth} >{\centering\arraybackslash}p{0.08\textwidth} >{\centering\arraybackslash}p{0.095\textwidth} >{\centering\arraybackslash}p{0.09\textwidth}@{}}
\toprule
\multirow{2}{*}{\textbf{Model}} & \multicolumn{4}{c}{\textbf{LLM-based Evaluation (Avg.)}} & \multicolumn{1}{c}{\textbf{Citation}} & \multicolumn{2}{c}{\textbf{Cost Efficiency}} \\
\cmidrule(lr){2-5} \cmidrule(lr){6-6} \cmidrule(l){7-8}
 & \textbf{Cov.}~\textcolor{green!60!black}{$\uparrow$} & \textbf{Logic}~\textcolor{green!60!black}{$\uparrow$} & \textbf{Rel.}~\textcolor{green!60!black}{$\uparrow$} & \textbf{Overall}\,\textcolor{green!60!black}{$\uparrow$} & \textbf{Ver.(\%)}\,\textcolor{green!60!black}{$\uparrow$} & \textbf{Tokens}\,\textcolor{red}{$\downarrow$} & \shortstack{\textbf{Inf. Time}\\\textbf{(s)}\,\textcolor{red}{$\downarrow$}} \\
\midrule
PRIMERA & 1.18 & 1.41 & 2.04 & 1.54 & 0.00 & \textbf{0.34K} & \textbf{2} \\
\midrule
GPT-4o (Long Context) & 3.20 & 3.28 & 3.76 & 3.41 & 48.61 & 6.61K & 38 \\
GPT-4o (RAG) & 3.19 & 3.30 & 3.76 & 3.42 & \underline{96.40} & \underline{3.68K} & \underline{37} \\
\midrule
SRW (Qwen 2.5 72B) & 3.38 & \textbf{3.73} & \underline{4.17} & 3.76 & 80.40 & 210.37K & 585 \\
Orchestrator (Qwen 2.5 72B) & \underline{3.41} & \underline{3.71} & \textbf{4.27} & \underline{3.80} & \textbf{99.67} & 181.29K & 822 \\
\midrule
\textbf{CREW (Qwen 2.5 72B, 4 Agents)} & \textbf{3.44} & \textbf{3.73} & \textbf{4.27} & \textbf{3.82} & \underline{99.33} & 198.74K & 755 \\
\bottomrule
\end{tabular*}
\vspace{0.8em}

\caption{Comparison with a naive orchestration strategy on the \textsc{OARelatedWork} dataset. The best and the runner-up results are shown in \textbf{bold} and \underline{underlined}, respectively.}
\label{tab:naive_orchestrator}
\begin{tabular*}{\textwidth}{@{\extracolsep{\fill}}>{\raggedright\arraybackslash}p{0.38\textwidth} >{\centering\arraybackslash}p{0.09\textwidth} >{\centering\arraybackslash}p{0.09\textwidth} >{\centering\arraybackslash}p{0.09\textwidth} >{\centering\arraybackslash}p{0.09\textwidth} >{\centering\arraybackslash}p{0.11\textwidth}@{}}
\toprule
\multirow{2}{*}{\textbf{Model}} & \multicolumn{4}{c}{\textbf{LLM-based Evaluation (Avg.)}} & \multicolumn{1}{c}{\textbf{Citation}} \\
\cmidrule(lr){2-5} \cmidrule(lr){6-6}
 & \textbf{Cov.}~\textcolor{green!60!black}{$\uparrow$} & \textbf{Logic}~\textcolor{green!60!black}{$\uparrow$} & \textbf{Rel.}~\textcolor{green!60!black}{$\uparrow$} & \textbf{Overall}\,\textcolor{green!60!black}{$\uparrow$} & \textbf{Ver.(\%)}\,\textcolor{green!60!black}{$\uparrow$} \\
\midrule
Naive Orchestrator (Qwen 2.5 72B) & 3.22 & 3.24 & 3.73 & 3.40 & 98.67 \\
Orchestrator (Qwen 2.5 72B) & \underline{3.41} & \underline{3.71} & \textbf{4.27} & \underline{3.80} & \textbf{99.67} \\
\midrule
\textbf{CREW (Qwen 2.5 72B, 4 Agents)} & \textbf{3.44} & \textbf{3.73} & \textbf{4.27} & \textbf{3.82} & \underline{99.33} \\
\bottomrule
\end{tabular*}
\end{table*}

\subsection{Comparison Results}
\label{sec:exp_main}
We present four primary experiments to validate our framework's capabilities, covering main performance, model transferability, benchmark transferability on a different dataset, and ablation analysis. The scalability analysis over different numbers of agents is deferred to Appendix~\ref{sec:appendix_e}.

\textbf{Main Results.} As shown in Table~\ref{tab:main_results}, CREW improves over the SRW baseline in overall LLM-judged quality by 0.06 points on the 1--5 scale and citation verification by 23.54\%, while reducing token usage by 12.90\%. These gains indicate that learned coordination improves generation quality, citation validity, and efficiency. Compared with our Orchestrator baseline, CREW achieves higher overall quality (+0.02 points on the 1--5 scale) and lower inference time. This modest quality margin is expected, as the Orchestrator inherits CREW's architecture, including the shared workspace, local knowledge, and action space, and differs primarily by replacing the learned decentralized policy with a centralized LLM coordinator, as described in Section~\ref{sec:exp_baselines}.
To examine this structural contribution, we compare against a Naive Orchestrator adapted from HuggingGPT~\citep{shen2023hugginggpt}, where the coordinator handles problem decomposition, task generation, and iterative task assignment based on prior agent feedback. Table~\ref{tab:naive_orchestrator} shows that this variant obtains Overall and Citation Verification scores of 3.40 and 98.67\%, respectively, substantially underperforming both the architecture-aware Orchestrator (3.80 Overall) and CREW (3.82 Overall). These results indicate that the strong performance of the Orchestrator derives substantially from CREW's structural design.
CREW further avoids the coordinator as a single point of failure. Under a simulated 5\% API failure rate, CREW and the Orchestrator encounter similar numbers of failed calls, but CREW finishes in 812 rather than 1012 seconds because unaffected agents can continue operating independently without depending on an orchestrator.

\begin{table*}[!t]
\centering
\small
\setlength{\tabcolsep}{4pt}
\caption{Model transferability across inference LLMs using the controller trained with Qwen 2.5 72B Instruct.}
\label{tab:transferability}
\begin{tabular*}{\textwidth}{@{\extracolsep{\fill}}>{\raggedright\arraybackslash}p{0.36\textwidth} >{\centering\arraybackslash}p{0.07\textwidth} >{\centering\arraybackslash}p{0.07\textwidth} >{\centering\arraybackslash}p{0.07\textwidth} >{\centering\arraybackslash}p{0.08\textwidth} >{\centering\arraybackslash}p{0.08\textwidth} >{\centering\arraybackslash}p{0.105\textwidth}@{}}
\toprule
\multirow{2}{*}{\textbf{Inference Model}} & \multicolumn{4}{c}{\textbf{LLM-based Evaluation (Avg.)}} & \multicolumn{1}{c}{\textbf{Citation}} & \multicolumn{1}{c}{\textbf{Cost Efficiency}} \\
\cmidrule(lr){2-5} \cmidrule(lr){6-6} \cmidrule(lr){7-7}
 & \textbf{Cov.}~\textcolor{green!60!black}{$\uparrow$} & \textbf{Logic}~\textcolor{green!60!black}{$\uparrow$} & \textbf{Rel.}~\textcolor{green!60!black}{$\uparrow$} & \textbf{Overall}\,\textcolor{green!60!black}{$\uparrow$} & \textbf{Ver.(\%)}\,\textcolor{green!60!black}{$\uparrow$} & \textbf{Tokens}\,\textcolor{red}{$\downarrow$} \\
\midrule
LLaMA 8B (Long-context) & 2.97 & 2.99 & 3.17 & 3.04 & 53.23 & 20.52K \\
LLaMA 8B + CREW & 3.31 & 3.42 & 3.96 & 3.57 & 82.01 & 161.49K \\
\midrule
Gemini 2.0 Flash (Long-context) & 3.11 & 3.17 & 3.54 & 3.28 & 98.61 & 20.93K \\
Gemini 2.0 Flash + CREW & 3.31 & 3.60 & 4.23 & 3.71 & 96.70 & 156.10K \\
\midrule
Qwen 2.5 72B Instruct (Long-context) & 3.19 & 3.35 & 3.73 & 3.42 & 65.83 & 21.57K \\
Qwen 2.5 72B Instruct + CREW & 3.31 & 3.33 & 4.17 & 3.79 & 96.20 & 121.81K \\
\midrule
GPT-4o (Long Context) & 3.20 & 3.28 & 3.76 & 3.42 & 48.61 & 6.61K \\
GPT-4o + CREW & 3.43 & 3.77 & 4.26 & 3.82 & 96.70 & 139.10K \\
\bottomrule
\end{tabular*}
\vspace{0.8em}

\caption{Cross-domain transferability on Multi-News without policy retraining or reward adaptation. Best and runner-up results are shown in \textbf{bold} and \underline{underlined}.}
\label{tab:benchmark_transferability}
\begin{tabular*}{\textwidth}{@{\extracolsep{\fill}}>{\raggedright\arraybackslash}p{0.28\textwidth} >{\centering\arraybackslash}p{0.085\textwidth} *{3}{>{\centering\arraybackslash}p{0.075\textwidth}} *{3}{>{\centering\arraybackslash}p{0.07\textwidth}}@{}}
\toprule
\multirow{2}{*}{\textbf{Framework}} & \multicolumn{4}{c}{\textbf{LLM-based Evaluation (Avg.)}} & \multicolumn{3}{c}{\textbf{ROUGE}} \\
\cmidrule(lr){2-5} \cmidrule(l){6-8}
& \textbf{Overall}~\textcolor{green!60!black}{$\uparrow$} & \textbf{Cov.}~\textcolor{green!60!black}{$\uparrow$} & \textbf{Logic}~\textcolor{green!60!black}{$\uparrow$} & \textbf{Rel.}~\textcolor{green!60!black}{$\uparrow$} & \textbf{R-1}~\textcolor{green!60!black}{$\uparrow$} & \textbf{R-2}~\textcolor{green!60!black}{$\uparrow$} & \textbf{R-L}~\textcolor{green!60!black}{$\uparrow$} \\
\midrule
GPT-4o (Long Context) & 3.53 & 3.36 & 3.70 & 3.54 & 40.57 & 11.56 & 19.24 \\
SRW (Qwen 2.5 72B) & 3.37 & 3.20 & 3.59 & 3.31 & 39.13 & 12.18 & 19.24 \\
Orchestrator (Qwen 2.5 72B) & \underline{3.75} & \underline{3.53} & \underline{3.76} & \underline{3.96} & \underline{41.81} & \underline{13.03} & \textbf{20.61} \\
\midrule
\textbf{CREW (Ours)} & \textbf{3.84} & \textbf{3.65} & \textbf{3.91} & \textbf{3.97} & \textbf{43.74} & \textbf{14.08} & \underline{20.29} \\
\bottomrule
\end{tabular*}
\vspace{0.8em}

\caption{Ablation study of key CREW components and their effects on generation quality and citation reliability.}
\label{tab:ablation}
\begin{tabular*}{\textwidth}{@{\extracolsep{\fill}}>{\raggedright\arraybackslash}p{0.36\textwidth} >{\centering\arraybackslash}p{0.07\textwidth} >{\centering\arraybackslash}p{0.07\textwidth} >{\centering\arraybackslash}p{0.07\textwidth} >{\centering\arraybackslash}p{0.08\textwidth} >{\centering\arraybackslash}p{0.08\textwidth} >{\centering\arraybackslash}p{0.105\textwidth}@{}}
\toprule
\multirow{2}{*}{\textbf{Component Configuration}} & \multicolumn{4}{c}{\textbf{LLM-based Evaluation (Avg.)}} & \multicolumn{1}{c}{\textbf{Citation}} & \multicolumn{1}{c}{\textbf{Cost Efficiency}} \\
\cmidrule(lr){2-5} \cmidrule(lr){6-6} \cmidrule(lr){7-7}
 & \textbf{Cov.}~\textcolor{green!60!black}{$\uparrow$} & \textbf{Logic}~\textcolor{green!60!black}{$\uparrow$} & \textbf{Rel.}~\textcolor{green!60!black}{$\uparrow$} & \textbf{Overall}\,\textcolor{green!60!black}{$\uparrow$} & \textbf{Ver.(\%)}\,\textcolor{green!60!black}{$\uparrow$} & \textbf{Tokens}\,\textcolor{red}{$\downarrow$} \\
\midrule
w/o Draft Quality Reward & 3.50 & 3.70 & 4.30 & 3.83 & 44.07 & 139.02K \\
w/o Feedback Action & 2.33 & 2.54 & 2.85 & 2.58 & 10.00 & 37.54K \\
w/o Social Activity Observation & 2.97 & 3.30 & 3.65 & 3.30 & 0.00 & 85.57K \\
\midrule
CREW (Full System) & 3.45 & 3.73 & 4.27 & 3.82 & 96.20 & 152.70K \\
\bottomrule
\end{tabular*}
\end{table*}

\textbf{Model Transferability.} We evaluate whether the learned 3-agent coordination policy transfers across backbones without retraining the controller. As shown in Table~\ref{tab:transferability}, the policy trained with Qwen 2.5 72B Instruct remains effective when agents use other inference models to execute selected actions. CREW raises the overall score of LLaMA 8B from 3.04 to 3.57 and Gemini 2.0 Flash from 3.28 to 3.71, while GPT-4o reaches 3.82. Although citation verification varies with the generation model, every transferred configuration substantially exceeds the corresponding long-context setting in overall quality. These results indicate that the controller learns a reusable coordination protocol rather than model-specific generation patterns. The consistent gains across model sizes and providers suggest that coordination knowledge is largely decoupled from the linguistic capabilities of the executor, allowing CREW to adopt upgraded backbones without repeating policy training.

\begin{figure*}[!t]
\centering
\includegraphics[width=\textwidth]{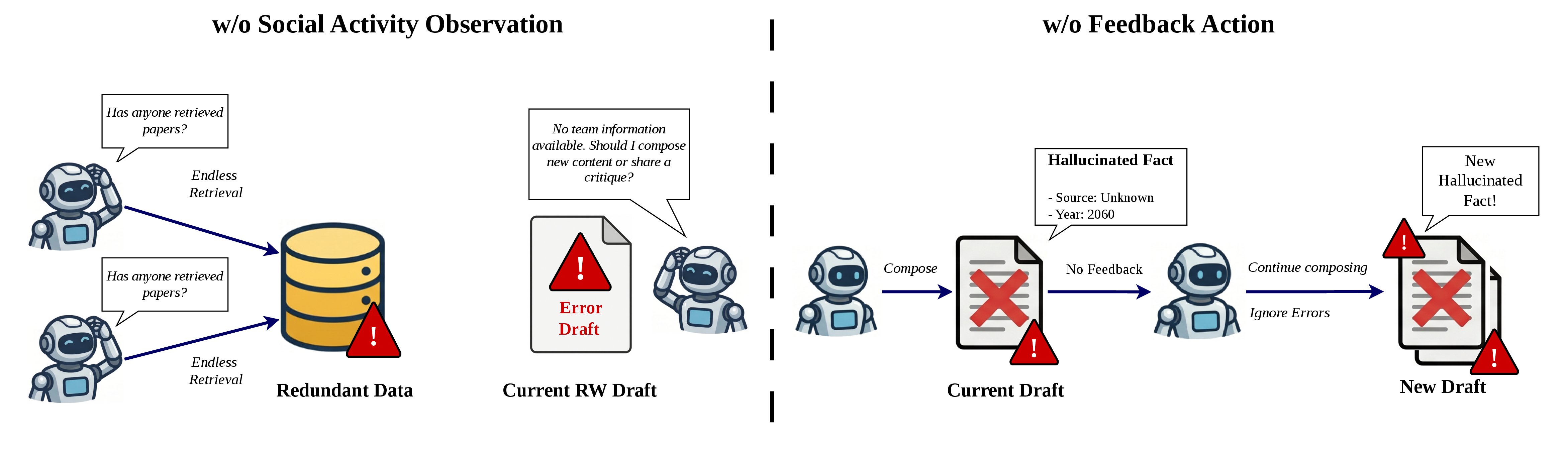}
\caption{Ablation failure cases of CREW. Removing social activity observation leads to poor coordination and redundant agent behavior, while removing the feedback action prevents agents from correcting citation errors introduced by others, reducing both citation accuracy and logical coherence.}
\label{fig:ablation_failure_cases}
\end{figure*}

\textbf{Benchmark Transferability.} We study cross-domain generalization on Multi-News~\citep{fabbri2019multi}, where the system must synthesize multiple news articles describing the same event. We adapt only the task prompts and represent articles for each event as a fully connected graph; the learned policy and reward-trained controller remain frozen. Alongside LLM-based quality dimensions, we report ROUGE-1, ROUGE-2, and ROUGE-L to measure lexical agreement with reference summaries. Table~\ref{tab:benchmark_transferability} shows that CREW achieves an overall score of 3.84, outperforming the Orchestrator and SRW by 0.09 and 0.47 points, respectively. It also obtains the best Coverage, Logic, Relevance, ROUGE-1, and ROUGE-2 scores, while remaining competitive on ROUGE-L. The gains across semantic judgments and reference-based metrics suggest that the policy preserves useful collaboration behaviors when the document genre, output purpose, and evidence structure differ from scientific related work generation. Notably, these improvements require neither controller retraining nor reward adaptation, indicating that the learned action-selection strategy remains useful when the evidence units shift from scientific papers to news articles. This result broadens CREW's potential applicability to other multi-document synthesis tasks requiring agents to coordinate evidence gathering, integration, and revision across heterogeneous sources.

\textbf{Ablation Studies.}
We conduct ablation studies to examine the contribution of key components under the IPPO objective. As shown in Table~\ref{tab:ablation}, removing observation, reward, or action-space components generally degrades performance. Removing the draft-quality reward weakens the direct signal for improving the generated draft, causing agents to rely more on retrieval and feedback while performing fewer effective writing updates. Consequently, citation verification drops to nearly half of the full CREW system, suggesting that retrieved evidence is less effectively grounded in the final draft. Although the overall score slightly increases in this setting, the gain is marginal and comes at the cost of substantially lower citation reliability.

Figure~\ref{fig:ablation_failure_cases} illustrates two representative failure cases. Without social activity observation, agents cannot track the action patterns of others, leading to redundant retrieval, overlapping behaviors, and weaker coordination; accordingly, Coverage, Logic, and Relevance decrease to around 3.30 on average. Removing the feedback action causes the largest degradation, with the Overall score dropping to 2.58, because agents lose an explicit mechanism for flagging hallucinated citations, missing evidence, and inconsistent claims. These results highlight the importance of both coordination-aware observation and feedback-driven revision in CREW. 

\label{sec::results}

\section{Conclusion}
We introduced CREW, a collaborative multi-agent reinforcement learning framework for automated related work generation. Instead of relying on a fixed pipeline, CREW allows multiple LLM agents to dynamically select among retrieval, knowledge dissemination, drafting, and critique actions according to a learned IPPO policy. Agents make decentralized decisions from their local observations while coordinating through shared knowledge, drafts, and feedback. This combination preserves agent autonomy without sacrificing the information exchange required for coherent scientific synthesis. Experiments on the \textsc{OARelatedWork} benchmark show that CREW improves generation quality and citation grounding over strong baselines while maintaining competitive efficiency. Comparisons with centralized and naive orchestrators indicate that effective collaboration depends on both the system structure and learned action selection. Ablations show that feedback, social-activity observations, and quality-oriented rewards are important for coordination, revision, and citation reliability. Transfer experiments demonstrate that the learned policy remains effective across different inference backbones and can adapt for broader multi-document synthesis without retraining. Scalability experiments show that additional agents can contribute complementary retrieval, writing, and review capacity, although gains must be balanced against inference cost and latency. Overall, CREW demonstrates that reinforcement-learned, decentralized coordination offers a flexible alternative to rigid LLM pipelines and provides a promising foundation for adaptive, citation-aware scientific writing assistants.

\section*{Limitations}
Although CREW achieves strong empirical performance, several limitations remain. 
First, the current system operates solely over a fixed candidate pool rather than dynamically querying live scholarly databases, potentially missing newly published papers or relevant background literature. 
Second, our automated reward signals and LLM-dependent critique may not fully capture nuanced scholarly conventions, theoretical framing, or field-specific writing styles. 
Third, while the learned policy can be generalized to broader multi-document synthesis tasks beyond related work generation simply by modifying action prompts, our evaluation breadth across other domains is currently confined to a preliminary study on Multi-News. 
Fourth, due to resource constraints, our human evaluation remains relatively limited in scale, despite exhibiting strong alignment with LLM-as-a-judge assessments. 
Finally, while parallel agents reduce wall-clock latency in principle, practical deployments are still constrained by backbone LLM reasoning capabilities, API rate limits, and hardware availability. 
Future work will integrate dynamic scholarly search engines, expand human evaluation, and explore broader scientific domains.

\section*{Ethical Considerations}

CREW is intended to assist researchers in drafting and organizing related work sections, not to replace human scholarly judgment. Because LLM-based systems can still generate unsupported claims, misinterpret cited work, or overstate relationships among papers, users should carefully verify all generated statements, citations, and comparisons before publication. The system should be used as a writing aid that accelerates evidence organization and revision, while final responsibility for accuracy, attribution, and intellectual contribution remains with the authors. There is also a risk that automated related work generation could encourage superficial literature review practices if users accept outputs without critical reading. To mitigate this risk, CREW emphasizes citation grounding and critique actions, but these safeguards do not eliminate the need for human review. In addition, evaluations based on LLM judges may reflect biases present in the judging models, including preferences for fluent writing or dominant research paradigms. We therefore recommend transparent reporting when such tools are used and encourage future work on stronger provenance tracking, uncertainty estimation, and human-centered evaluation protocols.

\begingroup\small
\bibliography{custom}
\endgroup

\appendix

\section{Algorithmic and Methodological Details}
This section provides algorithmic and methodological details for the proposed framework.
\subsection{Pseudocode of the Proposed Framework}
\label{sec:appendix_A}
Appendix~A.1 summarizes the full training, sampling, and inference procedure of CREW. Algorithm~\ref{alg:mars-rwg} shows how agents interact with shared artifacts over multiple rounds, collect rewards from artifact quality improvements, update the shared IPPO, and finally generate the related work draft through learned action sampling at inference time.

\begin{algorithm*}[t]
\caption{CREW with Independent PPO}
\label{alg:mars-rwg}
\small
\begin{algorithmic}[1]
\renewcommand{\algorithmicrequire}{\textbf{Input:}}
\renewcommand{\algorithmicensure}{\textbf{Output:}}

\REQUIRE Target paper $p$, paper database $\mathcal{P}$, citation graph $\mathcal{G}$, agents $\mathcal{A}=\{A_i\}_{i=1}^{M}$, horizon $T$
\ENSURE Final related-work draft $D_T$

\STATE Initialize shared actor model $\pi_\theta$ and shared critic model $V_\phi$.
\WHILE{$step < step_{max}$}
    \STATE \textit{// Loop through training steps}
    \STATE Initialize shared batch buffer $\mathcal{W}\leftarrow\emptyset$.
    \WHILE{$\mathcal{W}$ does not reach a predefined size}
        \STATE Create $M$ empty episode caches $E^1, \dots, E^M$
        \STATE Initialize shared states $D_0,G_0,F_0,C_0 \leftarrow \emptyset$
        \STATE \textit{// Reset the environment}
        \STATE Initialize local memory $L_i^0\leftarrow\emptyset$ for each $A_i$
        \FOR{$t=1$ to $T$}
            \FOR{each agent $A_i\in\mathcal{A}$ in parallel}
                \STATE Observe $o_t^i=\textsc{Observe}(D_{t-1},G_{t-1},F_{t-1},C_{t-1},L_{t-1}^{i})$
                \STATE Sample $a_t^i\sim\pi_{\theta}(\cdot\mid o_t^i)$, where $a_t^i\in\{Retrieve, Disseminate, Critique, Compose\}$
                \IF{$a_t^i=Retrieve$}
                    \STATE $(L_t^i,C_t)\leftarrow\textsc{Retrieve}(L_{t-1}^{i},C_{t-1},\mathcal{P},\mathcal{G})$
                \ELSIF{$a_i^t=Disseminate$}
                    \STATE $G_t\leftarrow\textsc{Disseminate}(L_t^i,G_{t-1})$
                \ELSIF{$a_i^t=Critique$}
                    \STATE $F_t\leftarrow\textsc{Critique}(D_{t-1},L_t^i,G_{t-1})$
                \ELSIF{$a_i^t=Compose$}
                    \STATE $D_t\leftarrow\textsc{Compose}(D_{t-1},G_{t-1},L_t^i,F_{t-1},C_{t-1})$
                \ENDIF
                \STATE $r_t^i\leftarrow\textsc{Reward}(D_t,G_t,F_t,C_t,L_i^t)$
                \STATE Store $(o_t^i,a_t^i,r_t^i,\log\pi_{\theta}(a_t^i\mid o_t^i),V_\phi(o_t^i))$ into cache $E^i$
            \ENDFOR
        \ENDFOR
        \FOR{each agent $A_i\in\mathcal{A}$}
            \STATE Append cache $E^i$ into the shared batch buffer $\mathcal{W}$.
        \ENDFOR
    \ENDWHILE
    
    \STATE \textit{// Stop sampling, start training}
    \STATE Estimate advantages $\hat{A}$ with GAE on the shared batch $\mathcal{W}$
    \STATE Compute actor loss $J_{\pi}(\theta)$ and critic loss $J_{V}(\phi)$ on the shared batch $\mathcal{W}$
    \STATE Update shared parameters $\theta$ and $\phi$ using Adam and gradient clipping
\ENDWHILE

\STATE \textbf{Inference:} At inference time, each agent samples its action from the learned shared policy, $a_t^i \sim \pi_{\theta}(\cdot \mid o_t^i)$, over $T$ rounds, and the final draft $D_T$ is returned.
\end{algorithmic}
\end{algorithm*}

\subsection{Reward Formulation}
\label{app:reward}
This appendix details our reward function, which defines component-wise quality improvement rewards for four artifacts: draft, local knowledge, global knowledge, and feedback. These rewards provide dense and interpretable learning signals that encourage agents to produce coherent, informative, and target-paper-relevant related work drafts. Draft-level improvement is measured against the human-written related work section, denoted as $e_{\mathrm{gold}}$.

\begingroup
\setlength{\parskip}{0pt}
\setlength{\abovedisplayskip}{3pt plus 1pt minus 1pt}
\setlength{\belowdisplayskip}{5pt plus 1pt minus 1pt}
\setlength{\abovedisplayshortskip}{2pt plus 1pt minus 1pt}
\setlength{\belowdisplayshortskip}{4pt plus 1pt minus 1pt}

\par\medskip
\noindent(i) \emph{\textbf{Draft quality}} $\widehat{Q}_D$ evaluates semantic relevance alongside a \emph{compression-density} score, using lossless compression as a lightweight proxy for textual regularity~\citep{li2004similarity}:
\begin{equation}
\label{eq:reward_draft}
\widehat{Q}_D = \gamma_{D,1}\,c(D,\mathrm{gold}) + \gamma_{D,2}\,(1 - C(D)/|D|),
\end{equation}
where $C(D)$ is the lossless compressed byte-length and $|D|$ is the original byte-length of the draft.
\par\medskip
\noindent(ii) \emph{\textbf{Local quality}} $\widehat{Q}_L$ balances relevance and topic density against a redundancy penalty:
\begin{equation}
\label{eq:reward_local}
\widehat{Q}_L = \gamma_{L,1}\,c(L_i,\mathrm{gold}) + \gamma_{L,2}\,c(L_i,\mathrm{tgt}) - \gamma_{L,3}\,\rho_i,
\end{equation}
where $c(L_i,\mathrm{tgt})$ evaluates how topically focused the readings are, and $\rho_i=\max_{h\in\mathcal{H}_i}c(L_i,h)$ penalizes semantic overlap with the agent's recent reading history~$\mathcal{H}_i$.
\par\medskip
\noindent(iii) \emph{\textbf{Global quality}} $\widehat{Q}_G$ rewards alignment with the gold standard, semantic novelty relative to prior shared knowledge $G_{t-1}$, and lexical distributional similarity to the gold standard:
\begin{multline}
\label{eq:reward_global}
\widehat{Q}_G = \gamma_{G,1}\,c(G,\mathrm{gold}) + \gamma_{G,2}\,(1-c(G,G_{t-1}))\\
 + \gamma_{G,3}\,(1-\mathrm{JS}(p_G\|p_{\mathrm{gold}})),
\end{multline}
where $\mathrm{JS}(\cdot\|\cdot)$ measures lexical coverage of the global knowledge against the gold standard, since lower distributional divergence indicates that $G$ covers a more similar set of unigram evidence~\citep{lin1991divergence}.
\par\medskip
\noindent(iv) \emph{\textbf{Feedback quality}} $\widehat{Q}_F$ measures whether the critique targets draft deficiencies, using the gap vector $\mathbf{g}=\mathbf{e}_{\mathrm{gold}}-\mathbf{e}_D$:
\begin{equation}
\label{eq:reward_feedback}
\widehat{Q}_F = c\!\left(\mathbf{e}_F,\,\mathbf{g}\right).
\end{equation}
\par

\endgroup

\subsection{Joint Action and Credit Assignment}
\label{app:credit_assignment}

\noindent\textbf{Joint Collaboration.} When multiple agents simultaneously perform actions that modify shared resources, namely Disseminate, Critique, and Compose, our framework employs a segment-level coordination mechanism. Each agent selects the text segment it is most confident in editing and reports a confidence score $c_i \in [0,1]$ in JSON format. Because agents have distinct observations shaped by their individual exploration trajectories, they naturally gravitate toward different segments, mirroring human collaboration in which authors refine different parts of a manuscript in parallel. The environment then merges these edits into a cohesive document. If multiple agents select the same segment, the agent with the highest confidence score $c_i$ is granted editing rights.

We formalize the likelihood of such overlap using the Birthday Problem and Stirling's approximation~\citep{feller1968introduction}. Under a strictly uniform random policy, which serves as a worst-case baseline where agents select shared actions with probability $p=3/4$ and choose segments uniformly, the unconditional probability of overlapping edits for $M$ agents across $K$ segments is approximated as:
\begin{equation}
\begin{aligned}
P(\mathrm{overlap}) \approx
\sum_{n=2}^{M}
&\left[1-\exp\!\left(-\frac{n(n-1)}{2K}\right)\right] \\
&{}\times \binom{M}{n}\left(\frac{3}{4}\right)^n
\left(\frac{1}{4}\right)^{M-n}.
\end{aligned}
\end{equation}
For example, with $M=3$ agents and $K=6$ segments, the overlap probability is only $\approx 25.78\%$, suggesting that parallel co-editing remains efficient with limited coordination overhead.

\noindent\textbf{Credit Assignment.} Once the joint draft is generated, it is evaluated to compute a global reward. To fairly distribute this shared reward among participating agents, we draw inspiration from the counterfactual multi-agent policy-gradient framework COMA~\citep{foerster2018counterfactual} and apply difference rewards. Formally, the difference reward for agent~$i$ is defined as:
\begin{equation}
D_i(z)=G(z)-G(z_{-i}),
\end{equation}
where $G(z)$ is the global evaluation score of the fully composed text $z$, and $G(z_{-i})$ is the score of the text synthesized without the contributions of agent~$i$.

\section{Dataset Construction and Citation Graph}
\label{sec:appendix_b}
\noindent
To ensure a rigorous and reproducible evaluation of the generated related work sections, this appendix provides detailed information on the processing of the \textsc{OARelatedWork} dataset.
\begin{figure*}[t]
    \includegraphics[width=1\textwidth]{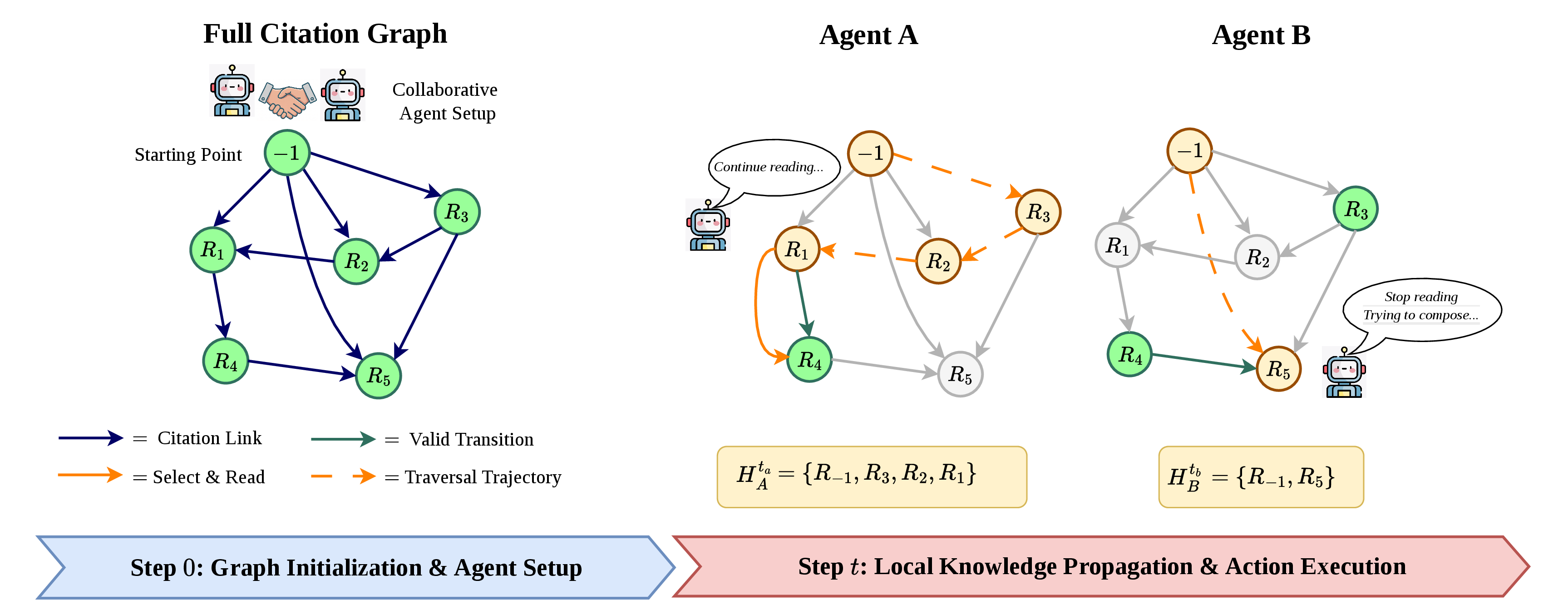}
    \caption{Citation graph construction and agent traversal in CREW. The target paper is assigned the sentinel ID $-1$ and linked to its cited references to form the working citation graph. Agents then traverse the graph under local observation, selecting reachable citations and incrementally updating their reading histories.}
    \label{fig:citation_graph}
\end{figure*}
\subsection{OARelatedWork Dataset}
\label{sec:oarelatedwork_dataset}

We leverage the \textsc{OARelatedWork} dataset, which contains 91{,}445 academic papers stored in JSONL format. Each paper record includes five fields: a unique integer \texttt{id}, \texttt{title}, \texttt{abstract}, \texttt{related\_work} (the gold-standard related work section), and \texttt{referenced} (the list of cited paper IDs).

In our experiments, we split the dataset into two non-overlapping subsets for policy learning and experimental evaluation. Specifically, we use 2\% of the papers for training and 0.4\% of the papers for evaluation. The training subset is used to optimize both the actor and critic networks during PPO training: the actor learns agents' action-selection policies, while the critic estimates value functions for evaluating state-action trajectories. The evaluation subset is kept disjoint from the training subset, ensuring that no target paper used for training appears in the test set.

\subsection{Citation Graph Construction}
\label{sec:graph_construction}

From the \textsc{OARelatedWork} dataset, we construct a directed citation graph $\mathcal{G} = (\mathcal{V}, \mathcal{E})$, as illustrated in Figure~\ref{fig:citation_graph}. The graph is initialized from the target paper, assigned the sentinel ID $-1$, and its connected reference papers; agents then traverse valid citation transitions to select and read papers, producing agent-specific trajectories and local histories. We represent the graph as two adjacency lists:
\begin{itemize}
    \item \textbf{Outgoing adjacency} (\texttt{citation\_adj\_out}): For each paper $p_i$, the list of papers it cites, i.e., $\mathcal{N}_{\text{out}}(p_i) = \{p_j \mid e_{i,j} \in \mathcal{E}\}$.
    \item \textbf{Incoming adjacency} (\texttt{citation\_adj\_in}): For each paper $p_i$, the list of papers that cite it, i.e., $\mathcal{N}_{\text{in}}(p_i) = \{p_j \mid e_{j,i} \in \mathcal{E}\}$.
\end{itemize}

During evaluation, a target paper $P_{\text{target}}$ is selected from the dataset and assigned the sentinel ID $-1$. Its outgoing citation edges are injected into the graph, and its direct neighbors (papers with at least 5 valid connections) serve as the reference pool $\mathcal{P} = \mathcal{N}_{\text{out}}(P_{\text{target}})$ from which the multi-agent system must read and synthesize. The gold-standard \texttt{related\_work} field of $P_{\text{target}}$ is used as the ground truth for evaluation.

\section{Evaluation metrics}
\label{sec:appendix_c}
\subsection{LLM-as-a-judge}
Evaluating long-form academic text generation is inherently complex, as traditional $n$-gram metrics fail to capture narrative flow, logical transitions, and citation discipline. We adopt an \textbf{LLM-as-a-Judge} protocol (using GPT-4o) scoring on a 1--5 Likert scale across three dimensions.

\textbf{Coverage.} Evaluates whether the generated related work comprehensively covers all key areas.
\begin{itemize}
    \setlength\itemsep{0em}
    \item \textit{Score 1:} Limited coverage, touching small portions and lacking key areas.
    \item \textit{Score 2:} Covers some parts but has noticeable omissions of significant areas.
    \item \textit{Score 3:} Generally comprehensive but misses a few key points.
    \item \textit{Score 4:} Covers most key areas comprehensively, with only minor topics left out.
    \item \textit{Score 5:} Comprehensively covers all key and peripheral topics with detailed discussions.
\end{itemize}
\textbf{Logic.} Assesses the logical flow and structural coherence.
\begin{itemize}
    \setlength\itemsep{0em}
    \item \textit{Score 1:} Lacks logic, no clear connections between sentences.
    \item \textit{Score 2:} Weak logical flow, content arranged in a disordered manner.
    \item \textit{Score 3:} Reasonable structure, though transitions could be improved (e.g., repeating).
    \item \textit{Score 4:} Good logical consistency, natural transitions, only slightly rigid.
    \item \textit{Score 5:} Tightly structured, logically clear, smooth transitions without redundancy.
\end{itemize}
\textbf{Relevance.} Evaluates focus on the target paper's core subject.
\begin{itemize}
    \setlength\itemsep{0em}
    \item \textit{Score 1:} Outdated or unrelated to the field, no alignment with the topic.
    \item \textit{Score 2:} Somewhat on topic but with several digressions.
    \item \textit{Score 3:} Generally on topic despite minor unrelated details.
    \item \textit{Score 4:} Mostly on topic and focused, without frequent digressions.
    \item \textit{Score 5:} Exceptionally focused, seamlessly contributing to core topic understanding.
\end{itemize}
\subsection{Citation Verification}

To complement LLM-based quality evaluation, we further assess whether the generated related work draft is grounded in the provided citation database. A high-quality draft should not only be coherent and relevant, but also cite the correct papers from the available reference pool without introducing unsupported or hallucinated references. We therefore use citation verification to measure the validity of citations produced in the generated draft.

Let $\hat{\mathcal{C}}(D)$ denote the set of paper IDs cited in a generated draft $D$, and let $\mathcal{P}_{\mathrm{ref}}$ denote the valid reference pool associated with the target paper. We define citation verification as the proportion of generated citations that can be correctly matched to valid papers in the reference pool:
\begin{equation}
\label{eq:citation_verification}
\mathrm{CV}(D)
=
\frac{
\left|\hat{\mathcal{C}}(D) \cap \mathcal{P}_{\mathrm{ref}} \right|
}{
\max\left(1, \left|\hat{\mathcal{C}}(D)\right|\right)
}.
\end{equation}
A higher $\mathrm{CV}(D)$ indicates that the generated draft cites papers grounded in the provided reference pool, while a lower score suggests the presence of invalid, unsupported, or hallucinated citations.
\section{Implementation Details}
\label{sec:appendix_d}
In this section, we provide the comprehensive setup details required to reproduce our framework, including system configurations, specific hyperparameter choices for IPPO, and the network architectures.
\subsection{System and Training Setup}
\label{sec:appendix_b1}
Our behavioral policy is optimized using IPPO, driven by the open-weight \textbf{Qwen 2.5 72B Instruct} model to ensure robust local reasoning during centralized training. 
The training setup involves a shared Actor and Critic network (Parameter Sharing) for multi agents interacting globally. 
The training is conducted for 3 training epochs, with $K=4$ PPO optimization epochs per update and an episode horizon (max rounds) of 15 steps per paper. Actions are capped to a context length of 6144 tokens, and GPU memory utilization is scaled dynamically.
For inference to demonstrate policy transferability, we swap the generation engine with LLaMA 8B, Gemini 2.0 Flash, Qwen 2.5 72B Instruct, and GPT-4o. Table~\ref{tab:system_parameters} summarizes the core IPPO hyperparameters, LLM settings, reward coefficients, and network configurations.

\subsection{Network Architectures}
\label{sec:appendix_b2}
The shared actor--critic layer configurations, including parameter counts, are detailed in Table~\ref{tab:system_parameters}. The observation vector contains 20 dimensions derived from graph features, draft quality, and social metrics.
\subsection{Baselines Detailed Analysis}
\label{sec:appendix_d3}
To accurately situate the performance of our framework within the current literature, we compare CREW against three groups of baseline paradigms: multi-agent methods, single-agent LLM methods, and abstractive supervised models.

\textbf{Multi-Agent Methods.} We include two multi-agent baselines. \textit{Select, Read, and Write} (SRW)~\citep{liu-etal-2025-select} is a specialized architecture for related-work generation that decomposes the task into selector, reader, and writer roles. For a fair comparison with our experimental budget, we run SRW for 15 rounds of select-read interactions, followed by one final write round to generate the related-work section. This setting preserves SRW's intended multi-stage evidence-gathering workflow while matching the number of interaction rounds used in our experiments. We also compare against an \textit{Orchestrator} baseline inspired by HuggingGPT~\citep{shen2023hugginggpt}. It keeps the same shared documents, agent pool, and action space as CREW but replaces the learned IPPO decision-making policy with a centralized LLM coordinator. The Orchestrator is run for the same 15 rounds as CREW and assigns actions to individual agents from the full system state. This baseline tests whether CREW's gains come from reinforcement-learned decentralized coordination rather than simply from using multiple agents and shared working memory.

\textbf{Single-Agent LLM Methods.} We evaluate GPT-4o under two single-agent settings. In \textit{GPT-4o (Long Context)}, the model receives the target paper context and the available reference information in a single prompt, then directly generates the related-work section. This setting represents a monolithic generation strategy that relies on the model's long-context reasoning ability. In \textit{GPT-4o (RAG)}, we first retrieve the most relevant chunks using embedding-based similarity and then prompt GPT-4o with the retrieved evidence, representing a standard retrieval-augmented search-and-generate pipeline.

\textbf{Abstractive Models.} PRIMERA~\citep{xiao-etal-2022-primera} is a supervised sequence-to-sequence model pre-trained for multi-document summarization. We adapt it to related-work generation on \textsc{OARelatedWork}, using it as an abstractive baseline that tests how far a specialized summarization model can go without explicit multi-agent reasoning, citation-graph exploration, or adaptive action selection.
\subsection{System Hyperparameters and Configurations}

We implemented CREW in Python~3.12.12 and used PettingZoo~\citep{terry2021pettingzoo} to construct the multi-agent reinforcement learning environment. All actor--critic components were trained locally on a workstation equipped with an NVIDIA RTX A5000 GPU with 24GB of GDDR6 memory. LLM-based generation and evaluation were conducted through the OpenRouter API~\citep{openrouter_api_2026}, which provides a unified interface for accessing different language model backends. Unless otherwise specified, all baselines and ablation studies were run under the same software and hardware environment. Table~\ref{tab:system_parameters} summarizes the system hyperparameters and configurations used in CREW, including IPPO optimization settings, LLM decoding parameters, and reward coefficients.

\begin{table}[!t]
\centering
\scriptsize
\setlength{\tabcolsep}{2pt}
\renewcommand{\arraystretch}{0.90}
\caption{System hyperparameters and configurations used in CREW.}
\label{tab:system_parameters}
\vspace{-0.6em}
\begin{tabular}{@{}p{0.13\columnwidth}p{0.14\columnwidth}p{0.35\columnwidth}ccc@{}}
\toprule
\textbf{Category} & \textbf{Notation} & \textbf{Description}
& \multicolumn{3}{c}{\textbf{Value}} \\
\midrule

\multirow{8}{*}{\textbf{RL}}
& $\alpha_{\omega}$ & Actor learning rate
& \multicolumn{3}{l}{$3\times10^{-4}$} \\
& $\alpha_{\phi}$ & Critic learning rate
& \multicolumn{3}{l}{$1\times10^{-3}$} \\
& $\gamma$ & Discount factor
& \multicolumn{3}{l}{$0.99$} \\
& $\lambda$ & GAE parameter
& \multicolumn{3}{l}{$0.95$} \\
& $\epsilon$ & Clipping threshold
& \multicolumn{3}{l}{$0.2$} \\
& $K$ & PPO epochs
& \multicolumn{3}{l}{$4$} \\
& $c_e$ & Entropy coefficient
& \multicolumn{3}{l}{$0.01$} \\
& $T$ & Rollout steps
& \multicolumn{3}{l}{$15$} \\

\midrule

\multirow{3}{*}{\textbf{LLM}}
& $\tau$ & Temperature
& \multicolumn{3}{l}{$0.3$} \\
& $p$ & Top-$p$ sampling
& \multicolumn{3}{l}{$0.9$} \\
& $N_{\max}$ & Max output tokens
& \multicolumn{3}{l}{$1400$} \\

\midrule

\multirow{8}{*}{\textbf{Reward}}
& $\gamma_{D,1}$ & Draft quality coefficient
& \multicolumn{3}{l}{$0.8$} \\
& $\gamma_{D,2}$ & Draft quality coefficient
& \multicolumn{3}{l}{$0.2$} \\
& $\gamma_{L,1}$ & Local quality coefficient
& \multicolumn{3}{l}{$0.5$} \\
& $\gamma_{L,2}$ & Local quality coefficient
& \multicolumn{3}{l}{$0.3$} \\
& $\gamma_{L,3}$ & Local quality coefficient
& \multicolumn{3}{l}{$0.2$} \\
& $\gamma_{G,1}$ & Global quality coefficient
& \multicolumn{3}{l}{$0.5$} \\
& $\gamma_{G,2}$ & Global quality coefficient
& \multicolumn{3}{l}{$0.3$} \\
& $\gamma_{G,3}$ & Global quality coefficient
& \multicolumn{3}{l}{$0.2$} \\

\midrule

\multirow{6}{*}{\textbf{Network}}
& $\pi_{1}$ & Policy hidden layer 1
& \multicolumn{3}{l}{$672$} \\
& $\pi_{2}$ & Policy hidden layer 2
& \multicolumn{3}{l}{$1056$} \\
& $\pi_{3}$ & Policy output layer 3
& \multicolumn{3}{l}{$132$} \\
\cmidrule(lr){2-6}
& $V_{1}$ & Value hidden layer 1
& \multicolumn{3}{l}{$672$} \\
& $V_{2}$ & Value hidden layer 2
& \multicolumn{3}{l}{$1056$} \\
& $V_{3}$ & Value output layer 3
& \multicolumn{3}{l}{$33$} \\

\bottomrule
\end{tabular}
\end{table}

\section{Scalability Analysis}
\label{sec:appendix_e}
\noindent\textbf{Multi-Agent Collaboration.}
To further analyze the effect of agent collaboration, we conduct an additional experiment by varying the number of agents from 1 to 5 while keeping the same backend language model for all settings. This ensures that the comparison reflects the contribution of the multi-agent organization rather than differences in model capability. For each configuration, we evaluate the generated related work quality, execution time, and total token consumption. As shown in Table~\ref{tab:multi_agent}, increasing the number of agents generally improves the quality scores, especially in logic and citation quality, since different agents can specialize in retrieval, feedback, knowledge updating, and writing. Meanwhile, execution time only increases moderately because agents can perform several actions in parallel, although total token usage grows with the number of agents. These results suggest that multi-agent collaboration provides a better trade-off between generation quality and computational cost than a single-agent setup.

\noindent\textbf{Discussion.}
Although parallel agents should ideally keep wall-clock time nearly stable, inference time increases mildly and then stabilizes in Table~\ref{tab:multi_agent}. This is largely due to OpenRouter API latency, which is external to our system. At timestep $t$, the system waits for the slowest agent, $T_t=\max_{1\leq i\leq M}\tau_i^{(t)}$. If each agent independently experiences a delayed API response with probability $p$~\citep{feller1968introduction}, and response time is $T_{\mathrm{wait}}$ under delay and $T_{\mathrm{normal}}$ otherwise, then $\mathbb{E}[T_t]=T_{\mathrm{wait}}-(T_{\mathrm{wait}}-T_{\mathrm{normal}})(1-p)^M$. Therefore, $\lim_{M\rightarrow\infty}\mathbb{E}[T_t]=T_{\mathrm{wait}}$ for $0<p\leq1$. Thus, API variability may raise latency, but the expected timestep duration remains bounded rather than growing unboundedly with the number of agents.

\begin{table*}[!t]
\centering
\captionsetup{hypcap=false}
\footnotesize
\setlength{\tabcolsep}{3.5pt}
\caption{Impact of scaling the number of collaborative agents on generation quality, citation grounding, and token consumption.}
\label{tab:multi_agent}
\begin{tabular}{>{\raggedright\arraybackslash}p{0.205\textwidth} >{\centering\arraybackslash}p{0.075\textwidth} >{\centering\arraybackslash}p{0.075\textwidth} >{\centering\arraybackslash}p{0.075\textwidth} >{\centering\arraybackslash}p{0.08\textwidth} >{\centering\arraybackslash}p{0.085\textwidth} >{\centering\arraybackslash}p{0.09\textwidth} >{\centering\arraybackslash}p{0.085\textwidth}}
\toprule
\multirow{2}{*}{\textbf{Number of Agents}} & \multicolumn{4}{c}{\textbf{LLM-based Evaluation (Avg.)}} & \multicolumn{1}{c}{\textbf{Citation}} & \multicolumn{2}{c}{\textbf{Cost Efficiency}} \\
\cmidrule(lr){2-5} \cmidrule(lr){6-6} \cmidrule(lr){7-8}
 & \textbf{Cov.}~\textcolor{green!60!black}{$\uparrow$} & \textbf{Logic}~\textcolor{green!60!black}{$\uparrow$} & \textbf{Rel.}~\textcolor{green!60!black}{$\uparrow$} & \textbf{Overall}\,\textcolor{green!60!black}{$\uparrow$} & \textbf{Ver.(\%)}\,\textcolor{green!60!black}{$\uparrow$} & \textbf{Tokens}\,\textcolor{red}{$\downarrow$} & \textbf{Time(s)}\,\textcolor{red}{$\downarrow$} \\
\midrule
1 & 3.04 & 3.36 & 3.82 & 3.41 & 86.10 & 42.26K & 359\\
2 & 3.41 & 3.69 & 4.27 & 3.79 & 100.00 & 90.64K & 778\\
3 & 3.31 & 3.33 & 4.17 & 3.79 & 96.20 & 121.81K & 644 \\
4 & 3.44 & 3.73 & 4.27 & 3.82 & 99.33 & 198.74K & 755 \\
5 & 3.48 & 3.77 & 4.36 & 3.87 & 99.17 & 246.49K & 833\\
\bottomrule
\end{tabular}
\end{table*}

\section{Qualitative Experiments}
\label{sec:appendix_f}
\subsection{Human Evaluation}
\label{sec:human_evaluation}
To complement the automatic evaluation, we conducted a human evaluation on 20 papers randomly selected from the Artificial Intelligence domain. Following the evaluation protocol of SRW~\citep{liu-etal-2025-select}, three graduate students performed pairwise comparisons between outputs generated by CREW and SRW. The evaluators assessed each pair in terms of coverage, logic, relevance, and overall quality. CREW achieved win rates of 53\% for coverage, 60\% for logic, and 57\% for relevance, yielding an average overall win rate of 56.7\%. These results indicate that human judges consistently preferred CREW to SRW across all evaluation dimensions. This human preference closely aligns with the automatic comparison in Table~\ref{tab:main_results}, jointly demonstrating the effectiveness of CREW's learned coordination in producing higher-quality related work; the qualitative example in Section~\ref{sec:case_study} further supports this conclusion by showing that CREW provides more citation-grounded synthesis and clearer connections between prior studies and the target paper than the baselines.

\subsection{Case Study}
\label{sec:case_study}
We use a challenging target paper on tweet wikification to illustrate how CREW handles short-text entity linking, collective inference, graph-based semi-supervision, and low-label training while grounding claims in the provided citation pool. Figure~\ref{fig:case_study} shows the target paper, and Figure~\ref{fig:case_comparison} compares outputs from CREW and the baselines.

\section{Prompt Templates}
\label{sec:appendix_prompts}
\beginpromptappendix
We present the detailed prompt templates that drive each agent action. The \textsc{Retrieve} action is further organized into two sequential phases: \textit{paper selection}, which selects candidate papers related to the target work, and \textit{paper reading}, which reads the selected papers and extracts useful evidence for later writing. Other prompts are designed following common prompt-engineering practices \cite{brown2020language, wei2022chain, yao2023react, white2023prompt, liu2023pretrain}, including role specification, explicit task instructions, step-by-step task decomposition, and output format constraints. These designs help reduce ambiguity and make the generated outputs more stable and easier to evaluate \cite{liu-etal-2025-select}.

\begin{figure*}[p]
\centering
\begingroup
\setlength{\abovecaptionskip}{1pt}
\setlength{\belowcaptionskip}{1pt}

\begin{tcolorbox}[
   enhanced,
   colback=orange!5,
   colframe=orange!65!black,
   boxrule=0.7pt,
   arc=2mm,
   width=0.96\textwidth,
   left=4pt,
   right=4pt,
   top=3pt,
   bottom=3pt,
   title=Target Paper for Qualitative Case Study,
   colbacktitle=orange!65!black,
   coltitle=white,
   fonttitle=\bfseries\small,
   toptitle=2pt,
   bottomtitle=2pt,
   fontupper=\small\linespread{0.90}\selectfont
]
\textbf{Title.} \textit{Collective Tweet Wikification based on Semi-supervised Graph Regularization}

\smallskip
\textbf{Abstract.} Wikification for tweets aims to automatically identify each concept mention in a tweet and link it to a concept referent in a knowledge base such as Wikipedia. Due to the shortness of a tweet, a collective inference model incorporating global evidence from multiple mentions and concepts is more appropriate than a non-collective approach that links each mention independently... To identify semantically related mentions for collective inference, it detects meta path-based semantic relations through social networks. Compared with a strong supervised baseline trained from 100\% labeled data, the proposed approach achieves comparable performance with 31\% labeled data and obtains a 5\% absolute F1 gain with 50\% labeled data.
\end{tcolorbox}
\caption{Qualitative case-study target paper. The framed title and abstract of the target paper.}
\label{fig:case_study}

\vspace{0.35em}

\resizebox{\textwidth}{!}{%
\begin{minipage}{1.07\textwidth}
\centering
\footnotesize
\tcbset{caseboxstyle/.style={enhanced,colback=orange!5,colframe=orange!65!black,colbacktitle=orange!65!black,coltitle=white,fonttitle=\bfseries,boxrule=0.8pt,arc=3mm,left=4pt,right=4pt,top=3pt,bottom=3pt,valign=top,fontupper=\footnotesize\linespread{0.92}\selectfont}}
\begin{tcbraster}[raster columns=2,raster equal height=all,raster column skip=0.025\textwidth,raster row skip=0.25em,raster left skip=0pt,raster right skip=0pt]
\begin{tcolorbox}[caseboxstyle,title={GPT-4o (RAG)}]
\redhl{The task of wikification for tweets, which involves linking concept mentions in tweets to corresponding entries in a knowledge base}, shares similarities with various challenges in entity linking and relation extraction... Distant supervision, as explored in studies [110538048, 26702637, 5884944], utilizes existing knowledge bases to generate training data by aligning known facts with text. \redhl{However, this approach often introduces noise due to incorrect assumptions about entity relationships, necessitating methods to mitigate this issue, such as the label-free distant supervision method proposed in [26702637]}... Overall, the intersection of distant supervision, collective inference, and context expansion techniques provides a robust framework for addressing the challenges of wikification in tweets.
\end{tcolorbox}
\begin{tcolorbox}[caseboxstyle,title={GPT-4o (Long Context)}]
... Local methods, such as those employed by \redhl{SemTag (Dill et al., 2003) and Wikify! (Mihalcea and Csomai, 2007)}, \redhl{focus on resolving each mention independently by considering the similarity between a mention and an entity}... Recent advancements have focused on semi-supervised and unsupervised learning techniques to mitigate the reliance on large labeled datasets. \redhl{For instance}, semi-supervised methods, such as those explored by \redhl{Druck et al. (2008) and Bellare et al. (2009)}, incorporate unlabeled ...In the domain of semi-supervised learning, \redhl{techniques such as graph-based label propagation (Wang and Cohen, 2007) and posterior regularization (Ganchev et al., 2010) have shown promise in effectively utilizing both labeled and unlabeled data}. These methods are particularly advantageous in scenarios where ...
\end{tcolorbox}
\begin{tcolorbox}[caseboxstyle,title={PRIMERA}]
\redhl{Collective Tweet Wikification based on Semi-supervised Graph Regularization.} Wikification for tweets aims to automatically identify each concept mention in a tweet and link it to a concept referent in a knowledge base, e.g., Wikipedia. In order to identify semantically related mentions for collective inference, \redhl{we detect meta path-based semantic relations through social networks}. Compared to the strong supervised baseline trained from 100\% labeled data, our proposed approach achieves comparable performance with 31\% labeled information and obtains 5\% absolute F1 gain with 50\% labelled data.
\end{tcolorbox}
\begin{tcolorbox}[caseboxstyle,title={Select, Read and Write (SRW)}]
...One notable approach is the use of global coherence, where the relationships between multiple mentions...For instance, the work in \redhl{ ``Efficient Collective Entity Linking with Stacking'' [5310325]} proposes a fast collective disambiguation method that combines local and global... Similarly, \redhl{"Local and Global Algorithms for Disambiguation to Wikipedia" [7610609] }analyzes the effectiveness of global...\redhl{In summary, the integration of global evidence, context expansion, semi-supervised learning, graph-based methods}, and collaborative ranking \redhl{represents a promising direction}...
\end{tcolorbox}
\end{tcbraster}

\vspace{0.3em}

\begin{minipage}[t]{\textwidth}
\begin{tcolorbox}[caseboxstyle,title={CREW}]
The field of entity linking, particularly in the context of short texts like tweets, has seen significant advancements through the integration of collective inference models and graph-theoretic methods.... \greenhl{Efficient collective entity linking, as proposed by [5310325]}, addresses the challenge of disambiguating entity mentions in text by leveraging a stacking approach that combines local and global evidence.... \greenhl{Similarly, [9854940] explores the use of graph-theoretic methods}, specifically centrality and regularization, to enhance class-instance label propagation.\greenhl{ The work in [7610609] further contributes to this area by comparing local and global algorithms for disambiguation to Wikipedia. This study highlights the importance of utilizing the Wikipedia link structure to ensure coherent} disambiguation, although it notes that local disambiguation remains a strong baseline. Our proposed semi-supervised graph regularization model builds upon these foundational works by incorporating both local and global evidence from multiple tweets through fine-grained relations, and by utilizing meta path-based semantic relations from social networks to identify semantically related mentions \greenhl{Our proposed semi-supervised graph regularization model builds upon these foundational works} by incorporating both local and global evidence....
\end{tcolorbox}
\end{minipage}
\end{minipage}%
}
\caption{Case study comparing related work generation outputs from GPT-4o (Long Context), GPT-4o (RAG), Primera, SRW, and CREW. \redhl{Red highlights mark problematic content such as unsupported citations, weak synthesis, target-abstract copying, or list-like summaries}. Green highlights show CREW's \greenhl{citation-grounded synthesis and its ability to connect prior works to the target paper.}}
\label{fig:case_comparison}
\endgroup
\end{figure*}

\begin{figure*}[!t]
\centering
\noindent\begin{minipage}{\textwidth}
\begin{promptbox}[promptcompact]{Retrieve Action Prompt: Phase 1 -- Paper Selection}
\textbf{System Prompt:} You are a research worker with excellent paper reading skills.

\vspace{0.45em}
\textbf{User Prompt:} You are helping write the Related Work section of a scientific paper. Your task is to choose the next paper section to read that adds the most new knowledge, not more of the same.

\vspace{0.45em}
\textbf{Inputs:} Citation graph: \promptplaceholder{graph}; current paper (id=$-1$ = paper being written): \promptplaceholder{paper\_details}; cited papers: \promptplaceholder{cited\_papers}; citing papers: \promptplaceholder{citing\_papers}; working memory: \promptplaceholder{working\_memory}; reading history: \promptplaceholder{reading\_history}; team global knowledge: \promptplaceholder{global\_knowledge}.

\vspace{0.45em}
\textbf{Mandatory Reasoning:} (1) Coverage audit: list research themes already covered across your memory and team global knowledge, or write ``Coverage: empty.'' (2) Gap identification: list 2--3 themes not yet covered, considering different method types, modalities, problem formulations, time periods, and contrastive approaches. (3) Candidate scoring: score unread candidates HIGH if their abstract addresses a gap and LOW if their themes are already covered. (4) Selection: pick the highest-scoring HIGH candidate; if all are LOW, choose the one least similar to global knowledge.

\vspace{0.45em}
\textbf{Hard Rules:} Never select a paper whose themes are more than 60\% covered in global knowledge. Never select a paper whose abstract overlaps more than 60\% with reading history. If you have read 2+ papers on the same technique, switch clusters. The section must exist exactly in the paper's structure list: \texttt{abstract} or \texttt{related\_work}. Never repeat a (paper\_id, section) pair from reading history. Prefer \texttt{related\_work} over \texttt{abstract} when available.

\vspace{0.45em}
\textbf{Output Format:} Return one JSON object only: \texttt{\{\{"id": <paper\_id>, "section": "<section\_name>", "rationale": "<which gap this fills and why it is not already in global knowledge>"\}\}}.
\end{promptbox}
\end{minipage}
\caption{Prompt template for the RETRIEVE action, Phase 1: Paper Selection.}
\label{fig:prompt_retrieve_selection}
\end{figure*}

\begin{figure*}[!t]
\centering
\noindent\begin{minipage}{\textwidth}
\begin{promptbox}[promptcompact]{Retrieve Action Prompt: Phase 2 -- Paper Reading}
\textbf{System Prompt:} You are a research worker with excellent paper reading skills.

\vspace{0.45em}
\textbf{User Prompt:} You are updating a short working memory ($\leq$4096 tokens) for writing the Related Work section.

\vspace{0.45em}
\textbf{Inputs:} Current paper: \promptplaceholder{cur\_paper}; cited papers: \promptplaceholder{cited\_paper}; just-read section: \promptplaceholder{section} of paper \promptplaceholder{paper\_id}; verbatim content: \promptplaceholder{content}; reading history: \promptplaceholder{reading\_history}; previous memory: \promptplaceholder{memory}; current draft: \promptplaceholder{draft}; global knowledge: \promptplaceholder{global\_knowledge}.

\vspace{0.45em}
\textbf{Novelty Gate:} Before writing, state in one sentence the primary contribution of paper \promptplaceholder{paper\_id}. Then check whether this contribution already appears in memory, draft, or global knowledge. If it is more than 60\% overlapping, use minimal mode; if genuinely new, use full extraction.

\vspace{0.45em}
\textbf{Extraction Rules:} Minimal mode extracts one distinctive sentence with [ID]. Full extraction extracts 2--4 sentences covering the new method, dataset, finding, contrast, or limitation. Preserve only claims grounded in the provided content and cite using the paper ID in square brackets.

\vspace{0.45em}
\textbf{Output Requirements:} Return the updated working memory as plain text. No JSON and no markdown.
\end{promptbox}
\end{minipage}
\caption{Prompt template for the RETRIEVE action, Phase 2: Paper Reading.}
\label{fig:prompt_retrieve_reading}
\end{figure*}

\begin{figure*}[!t]
\centering
\noindent\begin{minipage}{\textwidth}
\begin{promptbox}[promptcompact]{Critique Action Prompt}
\textbf{System Prompt:} You are a researcher worker with excellent knowledge synthesis skills, specializing in providing constructive academic feedback.

\vspace{0.45em}
\textbf{User Prompt:} Synthesize feedback insights from your Local Knowledge and integrate them into the existing Shared Feedback to create a unified, high-quality feedback reference for the Related Work draft.

\vspace{0.45em}
\textbf{Inputs:} Current draft: \promptplaceholder{draft}; local knowledge: \promptplaceholder{local\_knowledge}; global knowledge base: \promptplaceholder{global\_knowledge}; current shared feedback: \promptplaceholder{current\_feedback}.

\vspace{0.45em}
\textbf{Evaluation Dimensions:} If current feedback is empty, generate a comprehensive feedback outline evaluating coverage, logic, relevance, diversity, and citation quality. For each dimension, score 1--5 and provide specific edits. Diversity is critical: check whether multiple methodologies, paradigms, and viewpoints are represented, and flag over-reliance on one cluster or technique. Citation quality checks unsupported claims, citation dumping, invented IDs, and whether citations are anchored inline.

\vspace{0.45em}
\textbf{Synthesis Rules:} Merge overlapping critiques, lead with the highest-impact problems, make every critique actionable with a concrete fix or suggested rewrite, and resolve contradictory prior feedback by keeping the stricter version.

\vspace{0.45em}
\textbf{Output Requirements:} Return the updated synthesized feedback as plain text. No JSON and no markdown.
\end{promptbox}
\end{minipage}
\caption{Prompt template for the CRITIQUE action.}
\label{fig:prompt_critique}
\end{figure*}

\begin{figure*}[!t]
\centering
\noindent\begin{minipage}{\textwidth}
\begin{promptbox}[promptcompact]{Compose Action Prompt:}
\textbf{System Prompt:} You are a research worker with excellent paper writing skills.

\vspace{0.45em}
\textbf{User Prompt:} Finalize the Related Work section of a scientific paper.

\vspace{0.45em}
\textbf{Target Context:} Paper abstract: \promptplaceholder{abstract}; cited papers: \promptplaceholder{cited\_paper}.

\vspace{0.45em}
\textbf{Research Intelligence:} Local knowledge: \promptplaceholder{local\_knowledge}; global knowledge: \promptplaceholder{global\_knowledge}.

\vspace{0.45em}
\textbf{Collaboration:} Existing draft: \promptplaceholder{draft}; peer feedbacks: \promptplaceholder{feedbacks}.

\vspace{0.45em}
\textbf{Strict Instructions:} Group studies by theme rather than summarizing papers in isolation. Use only paper IDs from \promptplaceholder{cited\_paper}; never invent IDs or use numeric indices like [1]. Ensure every claim is supported by provided knowledge. Use formal academic tone and limit length to approximately 500--700 words. If a draft exists, refine it for consistency and feedback rather than rewriting unnecessarily. Every paragraph must start with a clear thematic topic sentence.

\vspace{0.45em}
\textbf{Output Format:} Return only a JSON object: \texttt{\{\{"related\_work": "..."\}\}}.
\end{promptbox}
\end{minipage}
\caption{Prompt template for the COMPOSE action.}
\label{fig:prompt_compose}
\end{figure*}

\begin{figure*}[!t]
\centering
\noindent\begin{minipage}{\textwidth}
\begin{promptbox}[promptcompact]{Disseminate Action Prompt}
\textbf{System Prompt:} You are a researcher worker with excellent knowledge synthesis skills, capable of organizing complex information into clear and logical insights.

\vspace{0.45em}
\textbf{User Prompt:} Synthesize new insights from your Local Cache and integrate them into the existing Shared Knowledge Base to create a unified, high-quality reference for the entire agent network.

\vspace{0.45em}
\textbf{Inputs:} Current shared knowledge base: \promptplaceholder{shared\_knowledge}; new information from local cache: \promptplaceholder{local\_cache}.

\vspace{0.45em}
\textbf{Quality Rules:} If shared knowledge is empty, first generate a professional thematic outline and then append synthesized content into the outline. Do not simply append; merge overlaps, enrich existing points, and place new information logically. Group related concepts, explain relationships, and highlight contrasts using integrative phrasing such as ``While [ID1] focuses on X, [ID2] addresses Y by ...''. Resolve contradictions when possible or document divergent views with citations.

\vspace{0.45em}
\textbf{Citation Rules:} Preserve citations using only paper IDs in square brackets. Cite an ID only when the claim is explicitly supported by Local Cache or current Shared Knowledge. Do not invent claims during synthesis. Citations must appear in the same sentence immediately after supported claims; do not place large citation lists at paragraph ends. Remove untraceable citations.

\vspace{0.45em}
\textbf{Output Requirements:} Return the updated Shared Knowledge as plain text. No JSON and no markdown.
\end{promptbox}
\end{minipage}
\caption{Prompt template for the DISSEMINATE action.}
\label{fig:prompt_disseminate}
\end{figure*}

\begin{figure*}[!t]
\centering
\noindent\begin{minipage}{\textwidth}
\begin{promptbox}[promptcompact]{Long Context Synthesis Prompt}
\textbf{System Prompt:} You are an expert researcher with excellent paper writing skills.

\vspace{0.45em}
\textbf{User Prompt:} You are writing the Related Work section of a scientific paper.

\vspace{0.45em}
\textbf{Target Context:} Paper abstract: \promptplaceholder{abstract}.

\vspace{0.45em}
\textbf{Neighbor Abstracts:} \promptplaceholder{neighbor\_abstracts}.

\vspace{0.45em}
\textbf{Strict Instructions:} Synthesize abstracts into a professional Related Work section. Group studies by theme and do not summarize papers in isolation. Use only paper IDs from the provided list in square brackets. Maintain a formal academic tone and limit length to approximately 600--800 words. Every paragraph must start with a clear thematic topic sentence.

\vspace{0.45em}
\textbf{Output Format:} Return only a JSON object: \texttt{\{\{"related\_work": "..."\}\}}.
\end{promptbox}
\end{minipage}
\caption{Prompt template for long-context synthesis.}
\label{fig:prompt_long_context}
\end{figure*}

\begin{figure*}[!t]
\centering
\noindent\begin{minipage}{\textwidth}
\begin{promptbox}[promptcompact]{Orchestrator Prompt}
\textbf{System Prompt:} You are a senior research coordinator managing a team of AI research agents that collaboratively write a Related Work section. At each round, assign exactly one action to each agent to coordinate the team efficiently.

\vspace{0.45em}
\textbf{Available Actions:} \textsc{Search} reads a new paper and updates local memory; \textsc{Update} synthesizes local memory into the shared global knowledge; \textsc{Write} drafts or refines the Related Work section; \textsc{Feedback} reviews the current draft and provides constructive revision suggestions.

\vspace{0.45em}
\textbf{Inputs:} Round index: \promptplaceholder{round\_num}/\promptplaceholder{max\_rounds}; target title: \promptplaceholder{paper\_title}; target abstract: \promptplaceholder{abstract}; current draft preview: \promptplaceholder{draft\_preview}; global knowledge preview: \promptplaceholder{global\_preview}; shared feedback preview: \promptplaceholder{feedback\_preview}; agent states: \promptplaceholder{agent\_states}; recent action history: \promptplaceholder{action\_history}.

\vspace{0.45em}
\textbf{Coordination Rules:} Promote action diversity across agents; prioritize \textsc{Search} and \textsc{Update} when knowledge is sparse; use \textsc{Write} once sufficient evidence has been gathered; alternate \textsc{Write} and \textsc{Feedback} when a draft exists; avoid assigning actions that are ineffective for an agent's current local memory; balance work so that all agents contribute.

\vspace{0.45em}
\textbf{Phase Guidance:} In early rounds, build coverage through searching and updating. In middle rounds, balance additional evidence gathering with writing. In late rounds, focus on writing and feedback to polish the draft.

\vspace{0.45em}
\textbf{Output Format:} Return only valid JSON with a brief \texttt{reasoning} field and an \texttt{assignments} object mapping each agent ID to \texttt{\{"action": "<SEARCH|UPDATE|WRITE|FEEDBACK>", "rationale": "<why>"\}}. Do not output any extra text.
\end{promptbox}
\end{minipage}
\caption{Prompt template for the centralized orchestrator baseline.}
\label{fig:prompt_orchestrator}
\end{figure*}

\FloatBarrier
\finishpromptappendix

\end{document}